\documentclass[10pt]{article} % For LaTeX2e
\usepackage[preprint]{tmlr}

\usepackage{amsmath,amsfonts,bm}

\def\eqref#1{equation~\ref{#1}}
\def\ceil#1{\lceil #1 \rceil}

\def\1{\bm{1}}

\DeclareMathAlphabet{\mathsfit}{\encodingdefault}{\sfdefault}{m}{sl}
\SetMathAlphabet{\mathsfit}{bold}{\encodingdefault}{\sfdefault}{bx}{n}

\usepackage{hyperref}
\usepackage{url}

\title{\aut: An Out-Of-The-Box Persistence-Based Clustering Algorithm}

\author{\name Marius Huber \email marius.huber@uzh.ch \\
      \addr Department of Computational Linguistics \\
        University of Z\"{u}rich
      \AND
      \name Sara Kali\v{s}nik \email skalisnik@psu.edu \\
      \addr Department of Mathematics \\
        Pennsylvania State University
      \AND
      \name Patrick Schnider \email patrick.schnider@inf.ethz.ch \\
      \addr Department of Mathematics and Computer Science \\
        University of Basel \\
        Department of Computer Science \\
        ETH Z\"{u}rich}

\def\month{10}  % Insert correct month for camera-ready version
\def\year{2025} % Insert correct year for camera-ready version
\def\openreview{\url{https://openreview.net/forum?id=Qd7H5mAbzV}} % Insert correct link to OpenReview for camera-ready version

\usepackage{amssymb}
\usepackage{amsmath}
\usepackage{amsthm}
\usepackage{mathrsfs}
\usepackage{verbatim}
\usepackage{lscape}
\usepackage{mathtools}
\usepackage{svg}
\usepackage{float}
\usepackage{enumitem}
\usepackage{booktabs}
\usepackage[justification=centering]{caption}
\usepackage{subcaption}
\usepackage{longtable}
\usepackage{xspace}
\usepackage[ruled]{algorithm2e}
\SetAlgoSkip{bigskip}
\SetKwInput{KwData}{Input}
\SetKwInput{KwResult}{Output}
\SetKwComment{Comment}{// }{}

\theoremstyle{plain}
\newtheorem{theorem}{Theorem}[section]

\newtheorem{remark}[theorem]{Remark}
\newtheorem{definition}[theorem]{Definition}

\newcommand{\uu}[1]{\underline{x}}

\newcommand{\ZZ}{\mathbb{Z}}

\newcommand{\RR}{\mathbb{R}}

\newcommand{\GG}{\mathcal{G}}

\newcommand{\PPP}{\mathbb{P}}

\newcommand{\DD}{\mathcal{D}}

\newcommand{\aut}{AuToMATo\xspace}
\newcommand{\tom}{ToMATo\xspace}

\newcommand{\bneck}[2]{W_{\infty}(#1,#2)}
\newcommand{\bneckfin}[2]{W_{\infty}^{\mathrm{fin}}(#1,#2)}
\newcommand{\dgm}{\mathrm{Dgm}}

\graphicspath{{figures/}}

\begin{document}

\maketitle

\begin{abstract}
We present \aut, a novel clustering algorithm based on persistent homology. While \aut is not parameter-free per se, we provide default choices for its parameters that make it into an out-of-the-box clustering algorithm that performs well across the board. \aut combines the existing \tom clustering algorithm with a bootstrapping procedure in order to separate significant peaks of an estimated density function from non-significant ones. We perform a thorough comparison of \aut (with its parameters fixed to their defaults) against many other state-of-the-art clustering algorithms. We find not only that \aut compares favorably against parameter-free clustering algorithms, but in many instances also significantly outperforms even the best selection of parameters for other algorithms. \aut is motivated by applications in topological data analysis, in particular the Mapper algorithm, where it is desirable to work with a clustering algorithm that does not need tuning of its parameters. Indeed, we provide evidence that \aut performs well when used with Mapper. Finally, we provide an open-source implementation of \aut in Python that is fully compatible with the standard \emph{scikit-learn} architecture.
\end{abstract}

% \tableofcontents

%%%%%%%%%%%%%%%%%%%%%%%%%%%%%%%%%%%%%%%%%%%%%%%%%%%%%%%%%%%%%%%%%%%%%%%%%%%%%%%%%%%%%%%%%%%%%%%%

\section{Introduction}\label{sec:introduction}

Clustering techniques play a central role in understanding and interpreting data in a variety of fields. The idea is to divide a heterogeneous group of objects into groups based on a notion of similarity.  This similarity is often measured with a distance or a metric on a data set. 
There exist many different clustering techniques~\citep{clustering_textbook1, clustering_textbook2}, including hierarchical, centroid-based and density-based techniques, as well as techniques arising from probabilistic generative models. Each of these methods is proficient at finding clusters of a particular nature. 
Many of the most commonly used clustering algorithms require a selection of parameters, a process which poses a considerable challenge when applying clustering to real-world problems.

In this work, we present and implement \aut (\emph{Automated Topological Mode Analysis Tool}), a novel clustering algorithm based on the topological clustering algorithm ToMATo~\citep{chazal2013persistence}. The latter summarizes the prominences of peaks of a density function in a so-called persistence diagram. The user then selects a prominence threshold \(\tau\) and retains all peaks whose prominence is above this threshold, which results in the final clustering.
A simple heuristic to select \(\tau\) is to sort the peaks by decreasing prominence, and to look for the largest gap between two consecutive prominence values~\citep{chazal2013persistence}. While yielding reasonable results in general, this procedure is not very robust to small changes in the prominence values.

A more robust and sophisticated method is to perform a bottleneck bootstrap on the persistence diagram produced by \tom, which is precisely what \aut does. That is, given a persistence diagram obtained by running \tom on a point cloud, \aut produces a confidence region for that diagram with respect to the bottleneck distance, which translates into a choice of \(\tau\) that determines the final clustering. While \aut is not parameter-free per se, we provide default choices that make it perform well across the board. Unless stated otherwise, \aut will henceforth refer to our algorithm with its parameters set to these defaults.
We experimentally analyze the clustering performance of \aut and we find that it not only outperforms parameter-free clustering algorithms, but often also even the best choice of hyperparameters for many parametric clustering algorithms.
Parameter-free algorithms building on \tom exist in the literature, for example, in~\citet{cotsakis2021implementing} the final clustering is determined by fitting a curve to the values of prominence, and in~\citet{outlier} significant values are separated from non-significant ones by adapting the process that produces the persistence diagrams.
Indeed, the former algorithm is one of those that \aut is shown to outperform.

We envision one important application of \aut to be to the \emph{Mapper} algorithm, introduced in~\citet{singh2007topological}. Mapper constructs a graph that captures the topological structure of a data set. It relies on many parameters, one of them being a clustering algorithm applied to various chunks of the data. Algorithms that depend heavily on a good choice of a tunable hyperparameter are generally not good candidates for usage with Mapper, as the best choice for the hyperparameter can vary significantly over the different chunks, and manually choosing a different hyperparameter for each may not be possible in practice. Thus, most choices of hyperparameter will generally perform badly on some of the subsets, leading to undesired results of Mapper. Thus, \aut can be seen as progress towards finding optimal parameters for Mapper, which is an active area of research~\citep{mapper_parameter_selection_carriere, mapper_parameter_selection_wang, mapper_parameter_selection_wang_2}.
Running examples for Mapper with \aut, we see that it is indeed a good choice for a clustering algorithm in this application when compared to parametric clustering algorithm such as DBSCAN.

%%%%%%%%%%%%%%%%%%%%%%%%%%%%%%%%%%%%%%%%%%%%%%%%%%%%%%%%%%%%%%%%%%%%%%%%%%%%%%%%%%%%%%%%%%%%%%%%

\section{Background}\label{sec:background}

\subsection {Persistence and the \tom clustering algorithm}\label{subsec:tomato}

Both \tom and \aut rely on the theory of persistence~\citep{Edelsbrunner2002, Zomorodian2005, carlsson2014} to quantify the prominence of peaks of (an estimate of) a density function, and to build a hierarchy of peaks. Given a topological space \(X\) equipped with a density function $f\colon X \to \RR_{\geq 0}$, the first step of persistence is to build a filtration from $X$.

\begin{definition}\label{def:suplevel_set_filtration}
Let $X$ be a topological space, and let $f\colon X \to \RR$ be continuous. The \textbf{superlevel set filtration} of $(X, f)$ is the family of superlevel sets \(\left\{X_{\geq t}\mid t\in\RR\right\}\), where \(X_{\geq t}\coloneqq f^{-1}\left([t, \infty)\right)\).
\end{definition}

In the following, we assume for ease of exposition that all local extrema of \(f\) have distinct values.
The idea underlying \tom is to track the evolution of (the number of) connected components of $X_{\geq t}$ as $t$ ranges from $+\infty$ to $-\infty$. In that process, the number of connected components of \(X_{\geq t}\) remains constant, unless \(t\) passes through the value of a local extremum of \(f\).
As \(t\) passes through the value of a local maximum, a new connected component is ``born'' and added to the superlevel set \(X_{\geq t}\).
Similarly, as \(t\) passes through the value of a local minimum, two connected components of \(X_{\geq t}\) are merged into one.
\tom builds a hierarchy of local maxima of \(f\) by declaring that, as two components get merged, the component corresponding to the local maximum with higher value absorbs the other one and persists, whereas the component corresponding to the local maximum with lower value ``dies''. Therefore, to each local maximum we associate a pair $(b,d)$ where $b$ denotes the birth and $d$ the death time, respectively.
The evolution of the connected components can be concisely recorded in a persistence diagram.

\begin{figure}[ht]
    \centering
    \includegraphics[width=0.8\textwidth]{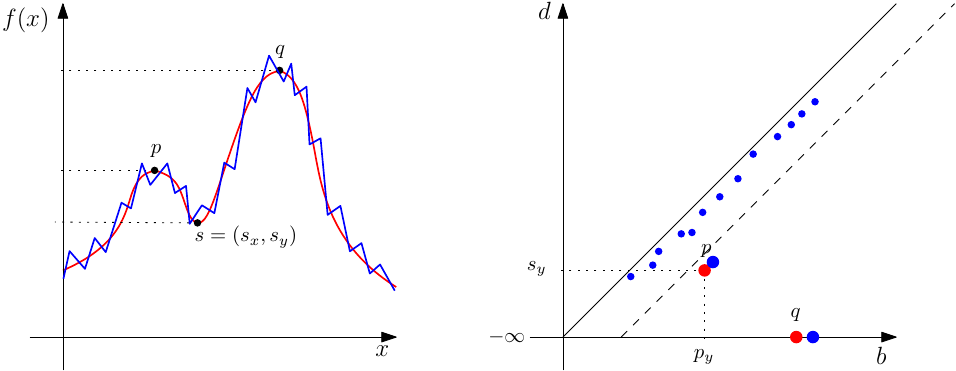}
    \caption{A function $f\colon K\to\RR$, \(K\subset\RR\), in red, and an estimate \(\hat{f}\) of $f$ in blue (left), with corresponding persistence diagrams \(\dgm(K, f)\) and \(\dgm(\GG, \hat{f})\) consisting of the red and  blue dots, respectively, together with a dashed line separating noise from features (right).}
    \label{fig:approx_persistence}
\end{figure}

\begin{definition}\label{def:persistence_diagram}
Let \(\{(b_{l}, d_{l})\}_{l}\) denote the birth and death times of connected components of the superlevel set filtration \(\left\{X_{\geq t}\right\}_{t\in\RR}\) associated to the density \(f\colon X \to \RR\).
The associated \textbf{persistence diagram}, denoted by \(\dgm(X,f)\), is the multiset in the extended plane \(\overline{\RR}^{2}\coloneqq\RR\cup\{\pm\infty\}\) consisting of the points \(\left\{(b_{l}, d_{l})\right\}_{l}\subset\overline{\RR}^{2}\) (counted with multiplicity) and the diagonal \(\Delta\coloneqq\left\{(x,x)\mid x\in\overline{\RR}\right\}\) (where each point on \(\Delta\) has infinite multiplicity).
For a given local maximum of \(f\) with birth time \(b_{l}\) and death time \(d_{l}\) , we refer to the difference \(d_{l}-b_{l}\) as its \textbf{prominence} or \textbf{lifetime}.
\end{definition}
The reason for working in the extended plane is that, provided that \(f\) has a global maximum, the superlevel set filtration \(X_{\geq t}\) will have a connected component that never dies, that is, has death time equal to \(-\infty\).
See the red graph in Figure~\ref{fig:approx_persistence} for an illustration.

The persistence diagram \(\dgm(X,f)\) provides a summary of \(f\). The points of \(\dgm(X,f)\) are in one-to-one correspondence with the local maxima of \(f\), and twice the $L^{\infty}$-distance of a point to the diagonal \(\Delta\) (that is, its Euclidean vertical distance) equals its prominence.

We now outline how the \tom clustering algorithm works. Given a point cloud \(X\) \tom relies on the assumption that the points of \(X\) were sampled according to some unknown density function \(f\). In a nutshell, \tom infers information about the local maxima of \(f\) by applying the above procedure to an estimate of \(f\). \tom takes as input:

\begin{itemize}[nolistsep]
\item {\bf A neighborhood graph \(\GG\) on the points of \(X\)}. Chazal et~al.\ mostly use the $\delta$-Rips graph and the \(k\)-nearest neighbor graph.\footnote{Given a point cloud, both of these undirected graphs have the set of data points as their vertex set. In the case of the \(\delta\)-Rips graph, two vertices are connected iff they are at a distance of at most \(\delta\) apart, whereas in the \(k\)-nearest neighbor graph, a data point is connected to another iff the latter is among the \(k\)-nearest neighbors of the first.}
\item {\bf A density estimator $\hat{f}$.} Each vertex $v$ of \(\GG\) is assigned a non-negative value $\hat{f}(v)$ that corresponds to the estimated density at $v$. Chazal et~al.\ propose two possible density estimators: the truncated Gaussian kernel density estimator and the distance-to-measure density, originally introduced in~\citet{biau:dtm_density}.\footnote{For a smoothing parameter \(m\in(0,1)\), and a given data point \(x\), its empirical (unnormalized) distance-to-measure density is given by \(\hat{f}(x)=\left(\frac{1}{k}\sum_{y\in N_{k}(x)}\Vert x-y\Vert^{2}\right)^{-\frac{1}{2}},
\) where \(k=\ceil{mn}\), \(N_{k}(x)\) denotes the set of the \(k\) nearest neighbors of \(x\), and \(n\) is the cardinality of the data set.}
\item {\bf A merging parameter $\tau\geq 0$.} This is a threshold that the prominence of a local maximum of the estimated density \(\hat{f}\) must clear for that local maximum to be deemed a feature.
\end{itemize}

Given the inputs above, \tom proceeds as follows.

\begin{enumerate}[nolistsep]
\item\label{item:tomato_1} {\bf Estimate the underlying density function $\hat{f}$ at the points of \(X\).}
\item\label{item:tomato_2} {\bf Apply a hill-climbing algorithm on \(\GG\).} Construct the neighborhood graph \(\GG\) on the points of \(X\), and construct a directed subgraph \(\GG'\) of \(\GG\) as follows: at each vertex $v$ of \(\GG\), place a directed edge from $v$ to its neighbor with highest value of $\hat{f}$, provided that that value is higher than $\hat{f}(v)$. If all neighbors of $v$ have lower values, $v$ is a peak of $\hat{f}$. This yields a collection of directed edges that form a spanning forest of the graph \(\GG\), consisting of one tree for each local maximum of \(\hat{f}\). In particular, these trees yield a partition of the elements of \(X\) into pairwise disjoint sets that serves as a candidate clustering on \(X\).
\item\label{item:tomato_3} {\bf Construct the persistence diagram.} Construct the persistence diagram \(\dgm(\GG, \hat{f})\) associated to the superlevel set filtration of \(\hat{f}\colon \GG\to\RR\).
\item\label{item:tomato_4} {\bf Merge non-significant clusters.} Iteratively merge every cluster of prominence less than \(\tau\) of the candidate clustering found in Step~\ref{item:tomato_2} into its parent cluster, that is, into the cluster corresponding to the local maximum that it gets merged into in the superlevel set filtration of \(\hat{f}\colon \GG\to\RR\). \tom outputs the resulting clustering of points of \(X\), in which every cluster has prominence at least \(\tau\) by construction.
\end{enumerate}

The reason why we can expect the persistence diagram of the approximated density to be ``close'' to the original one stems from the stability of persistence diagrams under the bottleneck distance (explained in Section \ref{subsec:bottleneck_bootstrap}). This is illustrated in Figure~\ref{fig:approx_persistence}.

In practice, the user must run \tom twice.
First, \tom is run with \(\tau=+\infty\) which is equivalent to computing the birth and death time of each local maximum of \(\hat{f}\) and hence the persistence diagram \(\dgm(\GG, \hat{f})\). From the diagram \(\dgm(\GG, \hat{f})\) the user then determines a merging parameter \(\tau\) by visually identifying a large gap in \(\dgm(\GG, \hat{f})\) separating, say, \(C\) points corresponding to highly prominent peaks from the rest of the points.
Then, \tom is run a second time with \(\tau\) set to that value, which results in the final clustering of \(X\) into \(C\) clusters.

\subsection{The bottleneck bootstrap}\label{subsec:bottleneck_bootstrap}

The bottleneck bootstrap, introduced in~\citet[Section 6]{chazal_bottleneck_bootstrap}, is used to separate significant features in persistence diagrams from non-significant ones. While it may be used in more general settings, we will restrict ourselves to the scenario of Section~\ref{subsec:tomato}.

We first review the bottleneck distance, which is the standard distance measure between persistence diagrams~\citep{EH10, chazal:hal-01330678}.
\begin{definition}
Let \(\dgm_{1}\) and \(\dgm_{2}\) be two persistence diagrams that have finitely many points off the diagonal. Let \(\pi\) denote the set of bijections \(\nu \colon\dgm_{1}\to\dgm_{2}\). Given points \(x=(x_{1}, x_{2})\) and \(y=(y_{1}, y_{2})\) in \(\overline{\RR}^{2}\), let \(\|x-y\|_\infty=\max\{|x_1-y_1|, |x_2-y_2|\}\) denote their \(L^{\infty}\)-distance, where we set \((+\infty)-(+\infty)=(-\infty)-(-\infty)=0\).
Then, the \textbf{bottleneck distance} between \(\dgm_{1}\) and \(\dgm_{2}\) is defined as \[
\bneck{\dgm_{1}}{\dgm_{2}} =\displaystyle\inf_{\nu\in\pi}\sup_{x\in\dgm_{1}} \|x-\nu(x)\|_\infty.
\]
\end{definition}
Note that a bijection \(\nu \colon\dgm_{1}\to\dgm_{2}\) is allowed to match an off-diagonal point of \(\dgm_{1}\) to the diagonal of \(\dgm_{2}\), and vice versa.

We now outline the bottleneck bootstrap. It relies on the following theorem, which summarizes the relevant results of~\citet[Section 6]{chazal_bottleneck_bootstrap}.

\begin{theorem}[\citet{chazal_bottleneck_bootstrap}]\label{thm:bottleneck_bootstrap}
    Let \(X\subseteq\RR^{N}\) be a sample consisting of \(n\) data points drawn according to a probability density function \(f\colon K\to[0,1]\), \(K\subset\RR^{N}\).
    Denote by \(\DD\coloneqq\dgm(K, f)\) and \(\widehat{\DD}\coloneqq\dgm(X, f)\) the corresponding unknown and estimated, respectively, persistence diagrams of superlevel sets.
    Given a confidence level \(\alpha\in(0,1)\), define \(q_{\alpha}\) by \[
        \PPP(\sqrt{n}\bneck{\DD}{\widehat{\DD}}\leq q_{\alpha})=1-\alpha.
    \]
    Then a consistent estimator for \(q_{\alpha}\) is given by \(\widehat{q}_{\alpha}\), which in turn is defined by\[
        \PPP(\sqrt{n}\bneck{\widehat{\DD}^{*}}{\widehat{\DD}}\leq \widehat{q}_{\alpha})=1-\alpha.
    \]
    Here, \(\widehat{\DD}^{*}\coloneqq\dgm(X^{*}, f)\) denotes the random persistence diagram constructed from a sample \(X^{*}\) of size \(n\) drawn according to the empirical measure \(P_{n}\) on \(X\), where \(P_{n}\) is defined as the probability measure on \(X\) that assigns the probability mass \(1/n\) to each data point in \(X\).
\end{theorem}

In our setting, the theorem above is applied as follows. Given the sample \(X\subseteq\RR^{N}\), we estimate \(f\) and the connectivity of \(K\) with a density estimator and a neighborhood graph, respectively (as explained in Section~\ref{subsec:tomato}).
This allows us to compute \(\widehat{\DD}\coloneqq\dgm(X, f)\), which, in turn, serves as an estimate of \(\DD\).
The empirical measure \(P_{n}\) on \(X\) serves as an approximation of the unknown probability measure \(f\), and using this, Theorem~\ref{thm:bottleneck_bootstrap} allows us to approximate the distribution \[
F(z)\coloneqq\PPP(\sqrt{n}\bneck{\DD}{\widehat{\DD}}\leq z)
\] with the distribution \[
\widehat{F}(z)\coloneqq\PPP(\sqrt{n}\bneck{\widehat{\DD}^{*}}{\widehat{\DD}}\leq z),
\] where \(\widehat{\DD}^{*}\coloneqq\dgm(X^{*}, f)\) is a random quantity.
Note that \(X^{*}\) may be thought of as a sample drawn from \(X\) with replacement.

Like the distribution \(F\), the distribution \(\widehat{F}\) is still not explicitly computable, but, unlike \(F\), it can be approximated by Monte Carlo as follows. We draw \(B\) samples \(X_{1}^{*},\dots,X_{B}^{*}\) of size \(n\) from \(P_{n}\), and for each of these \(B\) samples, we compute the persistence diagram \(\widehat{\DD}_{i}^{*}\coloneqq\dgm(X_{i}^{*}, f)\) and the quantity \(T_{i}^{*}\coloneqq\sqrt{n}\bneck{\widehat{\DD}_{i}^{*}}{\widehat{\DD}}\), \(i=1,\dots,B\).
Finally, we use the function \[
\widetilde{F}(z)\coloneqq\frac{1}{B}\sum_{i=1}^{B}\mathbf{1}_{[0, z]}(T_{i}^{*})
\] as an approximation of \(\widehat{F}\), and hence of \(F\).
Using this, we set \[
\widehat{q}_{\alpha}\coloneqq\inf\{z\mid \widetilde{F}(z)\geq 1-\alpha\}
\] to be our estimate of \(q_{\alpha}\).
This estimate is asymptotically consistent by Theorem~\ref{thm:bottleneck_bootstrap}, that is, \(\widehat{q}_{\alpha}\xrightarrow{n\to\infty}q_{\alpha}\).

In conclusion, the true, unknown persistence diagram \(\DD\) is at bottleneck distance of at most \(\widehat{q}_{\alpha}/\sqrt{n}\) from \(\widehat{\DD}\) with probability at least \(1-\alpha\). Hence, points of \(\widehat{\DD}\) that are at \(L^{\infty}\)-distance at most \(\widehat{q}_{\alpha}/\sqrt{n}\) from the diagonal could be matched to the diagonal under the bottleneck distance, and thus a point of \(\widehat{\DD}\) is declared to be a significant feature iff it is at \(L^{\infty}\)-distance of at least \(\widehat{q}_{\alpha}/\sqrt{n}\) to the diagonal, that is, iff its prominence is at least \(2\cdot\widehat{q}_{\alpha}/\sqrt{n}\).

%%%%%%%%%%%%%%%%%%%%%%%%%%%%%%%%%%%%%%%%%%%%%%%%%%%%%%%%%%%%%%%%%%%%%%%%%%%%%%%%%%%%%%%%%%%%%%%%

\section{Methodology and implementation of \aut}\label{sec:methodology_and_implementation}

\subsection{Methodology of \aut}\label{subsec:methodology}

\aut builds upon the \tom clustering scheme introduced in~\citet{chazal2013persistence} and implemented in~\citet{gudhi:PersistenceBasedClustering}.
\aut automates the step of visual inspection of the persistence diagram by means of the bottleneck bootstrap, thus promoting \tom to a clustering scheme that does not rely on human input.

More precisely, given a point cloud \(X\) to perform the clustering on, \aut takes as input

\begin{itemize}[nolistsep]
    \item an instance of \tom with fixed neighborhood graph and density function estimators;
    \item a confidence level \(\alpha\in(0,1)\); and
    \item a number of bootstrap iterations \(B\in\ZZ_{\geq 1}\).
\end{itemize}

\begin{remark}\label{rem:default_values}
    We point out that our implementation of \aut comes with default values for each of the objects.
    Each of these values can, of course, be adjusted by the user. For details on these default values, see Subsection~\ref{subsec:implementation}.
\end{remark}

\begin{figure}[ht]
    \centering
    \def\svgscale{1}
    %% Creator: Inkscape 1.4.2 (f4327f4, 2025-05-13), www.inkscape.org
%% Accompanies image file 'method.pdf' (pdf, eps, ps)
\begingroup%
  \makeatletter%
  \providecommand\color[2][]{%
    \errmessage{(Inkscape) Color is used for the text in Inkscape, but the package 'color.sty' is not loaded}%
    \renewcommand\color[2][]{}%
  }%
  \providecommand\transparent[1]{%
    \errmessage{(Inkscape) Transparency is used (non-zero) for the text in Inkscape, but the package 'transparent.sty' is not loaded}%
    \renewcommand\transparent[1]{}%
  }%
  \providecommand\rotatebox[2]{#2}%
  \newcommand*\fsize{\dimexpr\f@size pt\relax}%
  \newcommand*\lineheight[1]{\fontsize{\fsize}{#1\fsize}\selectfont}%
  \ifx\svgwidth\undefined%
    \setlength{\unitlength}{386.625bp}%
    \ifx\svgscale\undefined%
      \relax%
    \else%
      \setlength{\unitlength}{\unitlength * \real{\svgscale}}%
    \fi%
  \else%
    \setlength{\unitlength}{\svgwidth}%
  \fi%
  \global\let\svgwidth\undefined%
  \global\let\svgscale\undefined%
  \makeatother%
  \begin{picture}(1,0.3700291)%
    \lineheight{1}%
    \setlength\tabcolsep{0pt}%
    \put(0.12022105,0.27428866){\makebox(0,0)[t]{\lineheight{1.25}\smash{\begin{tabular}[t]{c}Data set $X$\end{tabular}}}}%
    \put(0.32201814,0.10740422){\makebox(0,0)[t]{\lineheight{1.25}\smash{\begin{tabular}[t]{c}Bootstrap samples\\$X_{1}^{*},\dots,X_{B}^{*}$\end{tabular}}}}%
    \put(0.41713326,0.30448101){\color[rgb]{0,0,0}\makebox(0,0)[t]{\lineheight{0}\smash{\begin{tabular}[t]{c}ToMATo PD\end{tabular}}}}%
    \put(0.60647823,0.11880111){\makebox(0,0)[t]{\lineheight{1.25}\smash{\begin{tabular}[t]{c}ToMATo PDs\end{tabular}}}}%
    \put(0,0){\includegraphics[width=\unitlength,page=1]{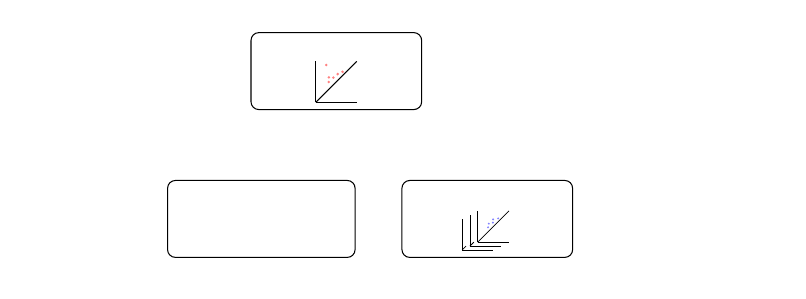}}%
    \put(0.82256488,0.30431026){\color[rgb]{0,0,0}\makebox(0,0)[t]{\lineheight{0}\smash{\begin{tabular}[t]{c}ToMATo PD with threshold\end{tabular}}}}%
    \put(0,0){\includegraphics[width=\unitlength,page=2]{method.pdf}}%
  \end{picture}%
\endgroup%

    \caption{Schematic of the methodology of \aut: from a data set \(X\), the usual \tom persistence diagram (with \(\tau=+\infty\)) is computed. Additionally, the analogous persistence diagrams are computed for the bootstrap samples \(X_{1}^{*},\dots,X_{B}^{*}\), which are created from \(X\) by drawing with replacement. Finally, the bootstrap procedure (indicated by \(\otimes\)) is used to compute a prominence threshold for the original persistence diagram.
    }
    \label{fig:method}
\end{figure}

To apply the bottleneck bootstrap as described in Section~\ref{subsec:bottleneck_bootstrap}, \aut generates \(B\) bootstrap subsamples \(X_{1}^{*},\dots,X_{B}^{*}\) of \(X\), each of the same cardinality as \(X\), where \(X\) is the data set whose points are to be clustered.
Then the underlying \tom instance with \(\tau=+\infty\) and its neighborhood graph and density function estimators is used to compute the persistence diagram for \(X\) and each of \(X_{1}^{*},\dots,X_{B}^{*}\), yielding persistence diagrams \(\widehat{\DD}\) and \(\widehat{\DD}_{1}^{*},\dots,\widehat{\DD}_{B}^{*}\), respectively. Using the bootstrapped diagrams \(\widehat{\DD}_{1}^{*},\dots,\widehat{\DD}_{B}^{*}\), a bottleneck bootstrap is performed on \(\widehat{\DD}\). This yields a value \(\widehat{q}_{\alpha}\) that (asymptotically as \(n\to\infty\)) satisfies \[
\PPP(\sqrt{n}\bneck{\DD}{\widehat{\DD}}\leq \widehat{q}_{\alpha})=1-\alpha,
\] where \(\DD\) denotes the persistence diagram of the true, unknown density function from which \(X\) was sampled. Thus, points of \(\widehat{\DD}\) of prominence at least \(2\cdot\widehat{q}_{\alpha}/\sqrt{n}\) are declared to be significant features of \(\widehat{\DD}\), and \aut outputs its underlying \tom instance with prominence threshold set to \(\tau=2\cdot\widehat{q}_{\alpha}/\sqrt{n}\).
This procedure is schematically depicted in Figure~\ref{fig:method}.

When computing the values \(\sqrt{n}\bneck{\widehat{\DD}_{i}^{*}}{\widehat{\DD}}\), \(i=1,\dots,B\), in the bottleneck bootstrap, we only consider points in \(\widehat{\DD}_{i}^{*}\) and \(\widehat{\DD}\) with finite lifetimes. The reason for this choice is that we consider peaks with infinite lifetime to be significant a priori. Moreover, some of the bootstrapped diagrams among the \(\widehat{\DD}_{1}^{*},\dots,\widehat{\DD}_{B}^{*}\) have a different number of points with infinite lifetime than the reference diagram \(\widehat{\DD}\). In these cases, the bottleneck distance of the bootstrapped diagram to the reference diagram is infinite, which heavily distorts the distribution \(\widetilde{F}(z)\).
This choice is justified by experiments.

\subsection{Implementation of \aut}\label{subsec:implementation}

We implemented \aut in Python, and all code with documentation is publicly available.\footnote{
    The code is archived on Zenodo (\href{https://doi.org/10.5281/zenodo.17279741}{\nolinkurl{doi.org/10.5281/zenodo.17279741}}) and developed openly on GitHub (\href{https://github.com/m-a-huber/automato_paper}{\nolinkurl{github.com/m-a-huber/automato\_paper}}).
}
For a description of \aut in pseudocode, see Algorithm~\ref{alg:automato}.
The algorithm has a worst-case complexity of \(O(B(nd+n\log(n)+N^{1.5}\log N))\), where \(d\) is the dimensionality of the data and \(N\) is the maximal number of off-diagonal points across all relevant persistence diagrams (which is generally much smaller than \(n\)); see below for details. Note that the factor of \(B\) can be significantly decreased through parallelization.

\begin{algorithm}[ht]
\caption{\aut}\label{alg:automato}
\KwData{point cloud \(X\) of \(n\) data points; instance \(\mathrm{tom}_{\tau}\) of \tom with neighborhood graph and density function estimators, and prominence threshold \(\tau\); confidence level \(\alpha\in(0,1)\); number of bootstrap iterations \(B\in\ZZ_{\geq 1}\).}
\BlankLine
\(\DD\gets \dgm(\mathrm{tom}_{\infty}(X))\) \Comment*[r]{compute persistence diagram of point cloud}
\For{\(i=1\) to \(B\)}{
    Let \(X_{i}^{*}\) be a subsample of \(X\) of size \(n\), sampled with replacement\;
    \(\DD_{i}^{*}\gets \dgm(\mathrm{tom}_{\infty}(X_{i}^{*}))\) \Comment*[r]{compute persistence diagram of subsample}
    \(d_{i}\gets\sqrt{n}\bneckfin{\DD_{i}^{*}}{\DD}\) \Comment*[r]{compute bottleneck distance between finite points}
}
Sort and reindex \(\{\DD_{1}^{*},\dots,\DD_{B}^{*}\}\) such that \(d_{1}\leq\cdots\leq d_{B}\)\;
\(k\gets \ceil{(1-\alpha)\cdot B}\)\;
\(\widehat{q}_{\alpha}\gets d_{k}\)\;
\(\tau\gets 2\cdot\widehat{q}_{\alpha}/\sqrt{n}\)\;
\BlankLine
\KwResult{\(\mathrm{tom}_{\tau}(X)\) \Comment*[r]{copy of initial \tom instance with prominence threshold set to \(\tau\)}}
\end{algorithm}

While the input parameters may be adjusted by the user, the implementation provides default values whose choices we discuss presently.

\paragraph{Choice of \tom parameters:} Our implementation of \aut is such that the user can directly pass parameters to the underlying \tom instance. If no such arguments are provided \aut uses the default choices for those parameters, as determined by the implementation of \tom given in~\citet{gudhi:PersistenceBasedClustering}. In particular, \aut uses the \(k\)-nearest neighbor graph and the (logarithm of the) distance-to-measure density estimators by default, each with \(k=10\).
Of course, the persistence diagrams produced by \tom, and hence the output of \aut, depend on this choice. This can lead to suboptimal clustering performance of \aut; see Section~\ref{sec:discussion}.

\paragraph{Choice of \(\alpha\) and \(B\):} By default, \aut performs the bootstrap on \(B=1000\) subsamples of the input point cloud, and sets the confidence level to \(\alpha=0.35\). The choice of this latter parameter means that \aut determines merely a 65\% confidence region for the persistence diagram produced by the underlying \tom instance. While in bootstrapping the confidence level is often set to, for instance,\(\alpha=0.05\), the seemingly strange choice of \(\alpha=0.35\) in the setting of \aut is justified by experiments. The value of 65\% seems to be low enough to offset some of the negative influence of using possibly non-optimized neighborhood graph and density estimators discussed in Section~\ref{sec:discussion}, while at the same time being high enough to yield good results when these estimators are chosen suitably. We point out that the value \(\alpha=0.35\) (as well as the value \(B=1000\)) was decided on after running an early implementation of \aut on just a few synthetic data sets. In particular, the choice was made \emph{before} conducting the experiments in Section~\ref{sec:experiments}. \aut is implemented in such a way that the parameter \(\alpha\) can be adjusted \emph{after} fitting and the clustering is automatically updated.

\paragraph{Complexity analysis of Algorithm~\ref{alg:automato}:} Recall from~\citet[Section 2]{chazal2013persistence} that, if an estimated density and a neighborhood graph are provided, \tom has a worst-case time complexity in \(O(n\log(n)+m\alpha(n))\), where \(n\) and \(m\) are the number of vertices and edges of the neighborhood graph, respectively, and \(\alpha\) denotes the inverse Ackermann function (note that \(n\) equals the number of data points).
By default, \tom (and hence \aut) works with the \(k\)-nearest neighbor graph and distance-to-measure density estimators, where the latter itself relies on the \(k\)-nearest neighbor graph (each with \(k=10\)). Taking into account the known complexity bound \(O(nd)\) for the creation of the \(k\)-nearest neighbor graph (where \(d\) is the dimensionality of the data), and using the fact that \(m\in O(n)\) for this graph, this leads to a worst-case time complexity in \(O(nd+n\log(n))\) for a single run of \tom. Creating the bootstrap samples \(X_{i}^{*}\), \(i=1,\dots,B\), has complexity in \(O(Bn)\); computing the values \(\sqrt{n}\bneckfin{\DD_{i}^{*}}{\DD}\), \(i=1,\dots,B\), has worst-case complexity \(O(BN^{1.5}\log(N))\) (where $N$ denotes the maximal number of off-diagonal points across all relevant persistence diagrams), and sorting them has worst-case complexity in \(O(B\log(B))\). Combined, this leads to a worst-case complexity for \aut in \(O(B(nd+n\log(n)+N^{1.5}\log(N))+B\log(B))\). Using that \(B\) is a constant, we obtain the runtime of \(O(B(nd+n\log(n)+N^{1.5}\log N))\) claimed above.

We point out that the complexity of \(O(N^{1.5}\log(N))\) for the computation of the bottleneck distance between a pair of persistence diagrams is a theoretical worst-case scenario, whose validity was established in~\cite{Efrat2001}. In practice, one typically allows for a small error in the computation of bottleneck distances which decreases the effective complexity~\citep{kerber_morozov_nigmetov_compare_persistence_diagrams}. Our implementation of \aut makes use of this by allowing for the smallest possible error as determined by the smallest positive double (see the implementation in ~\cite{gudhi:BottleneckDistance} for details).
Moreover, note that the factor \(B\) appearing in the complexity of \aut can be drastically decreased in practice through parallelization.

\paragraph{The Python package:} Our Python package for \aut consists of two separate modules; one for \aut itself, and one for the bottleneck bootstrap. Both are compatible with the \emph{scikit-learn} architecture, and the latter may also be used as a stand-alone module for other scenarios. In addition to the functionality inherited from the \emph{scikit-learn} API, the implementation of \aut comes with options of
\begin{itemize}[nolistsep]
    \item adjusting the parameter \(\alpha\) of a fitted instance of \aut which automatically updates the resulting clustering without repeating the (computationally expensive) bootstrapping;
    \item plotting the persistence diagram and the prominence threshold found in the bootstrapping;
    \item setting a seed in order to make the creation of the bootstrap subsamples in \aut deterministic, thus allowing for reproducible results; and
    \item parallelizing the bottleneck bootstrap for speed improvements.
\end{itemize}
Finally, our implementation of \aut contains a parameter that allows the algorithm to label points as outliers.
In a nutshell, a point is classified as an outlier if it is not among the nearest neighbors of more than a specified percentage of its own nearest neighbors. This feature, however, is currently experimental (and is thus turned off by default).

%%%%%%%%%%%%%%%%%%%%%%%%%%%%%%%%%%%%%%%%%%%%%%%%%%%%%%%%%%%%%%%%%%%%%%%%%%%%%%%%%%%%%%%%%%%%%%%%

\section{Experiments}\label{sec:experiments}

\subsection{Choice of clustering algorithms for comparison}\label{subsec:choice_of_algs_for_comparison}
We chose to compare \aut with its default parameters against
\begin{itemize}[nolistsep]
    \item DBSCAN and its extension HDBSCAN;
    \item hierarchical clustering with Ward, single, complete and average linkage;
    \item the FINCH clustering algorithm~\citep{sarfraz2019efficient}; and
    \item a clustering algorithm building on \tom stemming from the \emph{Topology ToolKit} (TTK) suite~\citep{topology_toolkit}; in the following, we will refer to this as the \emph{TTK-algorithm}.\footnote{For the \emph{Topology ToolKit}, see \href{https://topology-tool-kit.github.io/}{\nolinkurl{topology-tool-kit.github.io/}} (BSD license).}
\end{itemize}
For DBSCAN, HDBSCAN and the hierarchical clustering algorithms mentioned above, we worked with their implementations in \emph{scikit-learn}.\footnote{\href{https://scikit-learn.org/stable/modules/clustering.html}{\nolinkurl{scikit-learn.org/stable/modules/clustering.html}}} For the FINCH clustering algorithm, we worked with the version available on GitHub.\footnote{\href{https://github.com/ssarfraz/FINCH-Clustering}{\nolinkurl{github.com/ssarfraz/FINCH-Clustering}} (CC BY-NC-SA 4.0 license)} Indeed, we subclassed that version in order to make it compatible with the \emph{scikit-learn} API. Similarly, we created a \emph{scikit-learn} compatible version of the TTK-algorithm by combining code from TTK with the description of the algorithm given in~\citet[Section 5.2]{cotsakis2021implementing}. While we included DBSCAN and HDBSCAN among the clustering algorithms to compare \aut against because they are standard choices, we chose to include the hierarchical clustering algorithms because they are readily available through \emph{scikit-learn}. Finally, we chose to include FINCH and the TTK-algorithm because, like \aut, they are out-of-the-box (indeed, parameter-free) methods and are thus especially interesting to compare \aut against.

\subsection{Choice of data sets}\label{subsec:choice_of_data_sets_for_comparison}
The data sets on which we ran \aut and the above clustering algorithms stem from the \emph{Clustering Benchmarks} suite~\citep{gagolewski_benchmarking_suite}.\footnote{Specifically, we worked with version 1.1.0 of the benchmarking suite~\citep{gagolewski_benchmarking_suite_release}}
We chose this collection as it comes with a large variety of different data sets, all of which are labeled by one or more ground truths, allowing for a fair and extensive comparison. The collection contains five recommended batteries of data sets from which we selected those (data set, ground truth)-pairs that we deemed reasonable for a general purpose parameter-free clustering algorithm.
For instance, we chose to include the data set named \texttt{windows} that is part of the \texttt{wut}-battery, but not the data set named \texttt{windows} from the same battery (see Figure~\ref{fig:wut_and_olympic} in the appendix for an illustration).
We chose to include the \texttt{windows} data set because \aut determines clusters depending on connectivity, and topologically speaking, there is only one connected component in the \texttt{olympic} data set.
Finally, we excluded all instances where the ground truth contains data points that are labeled as outliers, as outliers creation is currently an experimental feature in \aut.

\subsection{Methodology of the experiments}\label{subsec:methodology_of_experiments}

We min-max scaled each data set, fitted the clustering algorithms to them, and recorded the clustering performance of each result by computing the Fowlkes-Mallows score~\citep{fowlkes1983method} of the clustering obtained and the respective ground truth.
While the Fowlkes-Mallows score was originally defined for hierarchical clusterings only, it may be defined for general clusterings as follows. Given a clustering \(C\) found by an algorithm and a ground truth clustering \(G\), one defines the Fowlkes-Mallows score as \[
    \mathrm{FMS} \coloneqq \sqrt{\frac{\mathrm{TP}}{\mathrm{TP} + \mathrm{FP}}}\cdot\sqrt{\frac{\mathrm{TP}}{\mathrm{TP} + \mathrm{FN}}},
\]
where 
\begin{itemize}
    \item \(\mathrm{TP}\) is the number of pairs of data points which are in the same cluster in \(C\) and in \(G\);
    \item \(\mathrm{FP}\) is the number of pairs of data points which are in the same cluster in \(G\) but not in \(C\); and
    \item \(\mathrm{FN}\) is the number of pairs of data points which are not in the same cluster in \(G\) but are in the same cluster in \(C\).
\end{itemize}
In other words, the Fowlkes-Mallows score is defined as the geometric mean of precision and recall of a classifier whose relevant elements are pairs of points that belong to the same cluster in both \(C\) and \(G\). It may attain any value between 0 and 1, and these extremal values correspond to the worst and best possible clustering, respectively.
We chose to use the Fowlkes-Mallows score as opposed to, for instance, mutual information or any of the Rand indices, because the latter have been shown to exhibit biased behavior depending on whether the clusters in the ground truth are mostly of similar sizes or not, see, for instance,~\citet{adjusting_for_chance_romano}; to the best of our knowledge, the Fowlkes-Mallows score does not suffer from such drawbacks.
Moreover, we chose not to use any intrinsic measures of clustering performance since any such measure implicitly defines a further clustering algorithm to compare \aut against, whereas we are interested in comparing \aut against a predefined ground truth clustering.

We set the hyperparameters of the HDBSCAN, FINCH and the TTK-algorithm to their default values (as per their respective implementations). In contrast to this, we let the distance threshold parameter for the DBSCAN and the hierarchical clustering algorithms vary from 0.05 to 1.00 in increments of 0.05, with the goal of comparing \aut against the best and worst performing version of these clustering algorithms. To account for the randomized component of \aut, we ran it ten times, each time with a different seed.

While we restricted ourselves to instances where the ground truth does not contain any points labeled as outliers, some of the clustering algorithms in our list (DBSCAN and HDBSCAN) label some data points as outliers. In order to prevent these algorithms from getting systematically low Fowlkes-Mallows scores because of these outliers, we removed all the points labeled as outliers by these algorithms, and only computed the Fowlkes-Mallows score on the remaining points, both for these clustering algorithms and for \aut. This of course gives an advantage to DBSCAN and HDBSCAN over \aut.

In order to allow reproducibility, we chose a fixed seed for all our experiments, which can be found in our code.\footnote{
    The code is archived on Zenodo (\href{https://doi.org/10.5281/zenodo.17279741}{\nolinkurl{doi.org/10.5281/zenodo.17279741}}) and developed openly on GitHub (\href{https://github.com/m-a-huber/automato_paper}{\nolinkurl{github.com/m-a-huber/automato\_paper}}).
}
We ran our experiments on a laptop with a 12th Gen Intel Core i7-1260P processor running at 2.10GHz.

\subsection{Results and interpretation}\label{subsec:results_and_comparison}

Table~\ref{table:summary_of_results} shows the average Fowlkes-Mallows score of each algorithm across all benchmarking data sets; for \aut, it shows the average and the standard deviation across the ten runs. For those benchmarking data sets that come with more than one ground truth, we included only the best score of the respective algorithm.
Similarly, we included only the best performing parameter selection for those algorithms that we ran with varying distance thresholds (which, of course, skews the comparison in favor of those algorithms).
As Table~\ref{table:summary_of_results} shows, \aut outperforms each clustering algorithm on average across all data sets, thus showing that it is indeed a versatile and powerful out-of-the-box clustering algorithm.
In particular, \aut outperforms the TTK-algorithm, which also build on \tom.

\begin{longtable}{lr}
\caption{Average clustering performance of AuToMATo vs. reference clustering algorithms}
\label{table:summary_of_results} \\
\toprule
Algorithm & Fowlkes-Mallows score \\
\midrule
\endfirsthead
\caption[]{Average clustering performance of AuToMATo vs. reference clustering algorithms} \\
\toprule
Algorithm & Fowlkes-Mallows score \\
\midrule
\endhead
\midrule
\multicolumn{2}{r}{Continued on next page} \\
\midrule
\endfoot
\bottomrule
\endlastfoot
AuToMATo & \textbf{0.8554±0.0228} \\
DBSCAN & 0.8457 \\
Average linkage & 0.8321 \\
HDBSCAN & 0.8209 \\
Single linkage & 0.8156 \\
TTK clustering algorithm & 0.8019 \\
Complete linkage & 0.7592 \\
Ward linkage & 0.5896 \\
FINCH & 0.5074 \\
\end{longtable}

\begin{figure}[ht]
    \centering
    \includesvg[inkscapelatex=false, width=\textwidth]{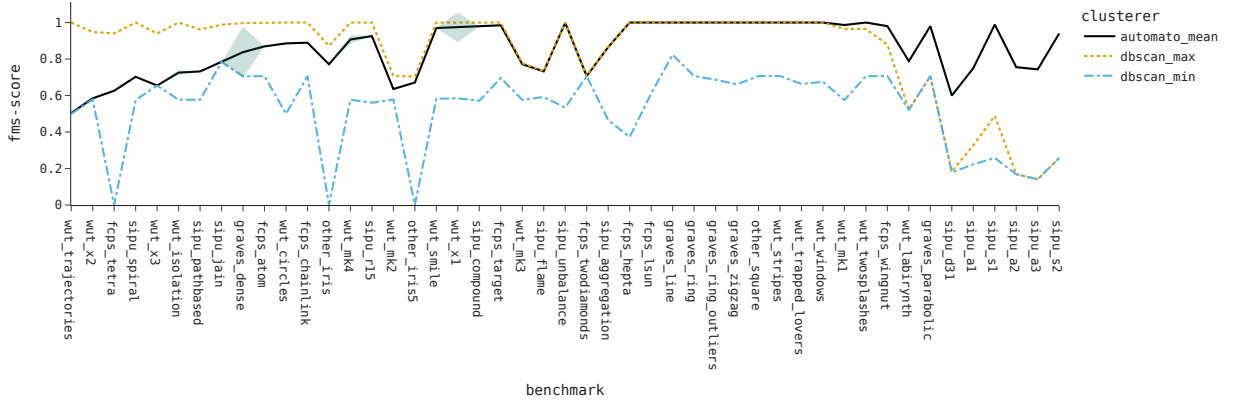}
    \caption{Fowlkes-Mallows score of \aut and DBSCAN across benchmarking data sets. The shading of ``automato\_mean'' indicates the standard deviation of the score across the ten runs.}
    \label{fig:aut_vs_dbscan_mainbody}
\end{figure}

The scores of our experiments are reported in Tables~\ref{table:aut_vs_dbscan} through~\ref{table:aut_vs_ttk} in Appendix~\ref{appendix_subsec:benchmarking_results}.
As an illustration, Figure~\ref{fig:aut_vs_dbscan_mainbody} shows that the best choice of parameter for DBSCAN sometimes outperforms \aut, which is to be expected. However, on most data sets where this is the case, the results from \aut are still competitive, and there is a significant number of instances where \aut outperforms DBSCAN for all parameter selections, in some cases by a lot.

%%%%%%%%%%%%%%%%%%%%%%%%%%%%%%%%%%%%%%%%%%%%%%%%%%%%%%%%%%%%%%%%%%%%%%%%%%%%%%%%%%%%%%%%%%%%%%%%

\section{Applications of \aut in combination with Mapper}\label{sec:sample_application_mapper}

The goal of \emph{Mapper} \citep{singh2007topological} is to approximate the \emph{Reeb graph} of a manifold $M$ based on a sample from $M$. The input is a point cloud $P$ with a filter function $P\rightarrow\RR$; a collection of overlapping intervals $\mathcal{U}=\{U_1,\ldots,U_n\}$ covering $\RR$; and a clustering algorithm. For each $U_i\in\mathcal{U}$, Mapper runs the clustering algorithm on the data points in the preimage $f^{-1}(U_i)$, creating a vertex for each cluster. Two vertices are then connected by an edge if the corresponding clusters (in different preimages) have some data points in common, yielding a graph that represents the shape of the data set.

\begin{figure}[ht]
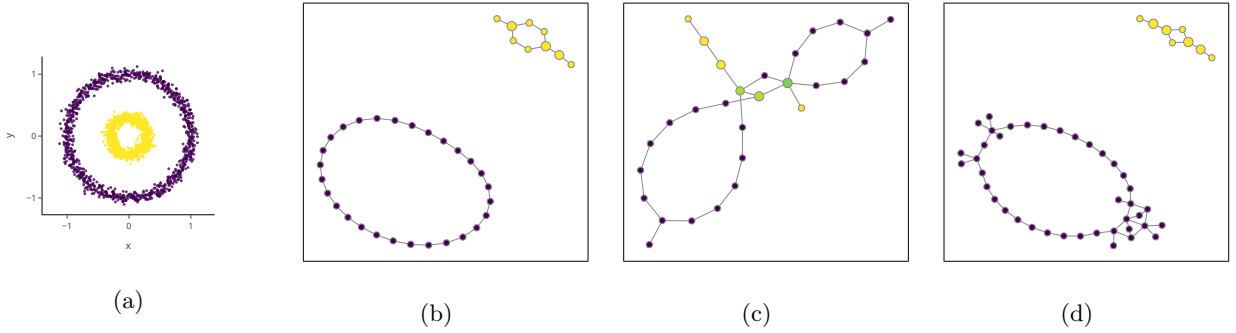

    \centering
    \begin{subfigure}[c]{0.23\textwidth}
        \centering
        \includesvg[inkscapelatex=false, width=\textwidth]{concentric_circles}
        \caption{}
        \label{fig:concentric_circles}
    \end{subfigure}
    \hfill
    \begin{subfigure}[c]{0.25\textwidth}
        \centering
        \includesvg[inkscapelatex=false, width=\textwidth]{mapper_concentric_circles_automato_15_intervals_0.3_overlap}
        \caption{}
        \label{fig:mapper_concentric_circles_automato}
    \end{subfigure}
    \hfill
    \begin{subfigure}[c]{0.25\textwidth}
        \centering
        \includesvg[inkscapelatex=false, width=\textwidth]{mapper_concentric_circles_dbscan_15_intervals_0.3_overlap}
        \caption{}
        \label{fig:mapper_concentric_circles_dbscan}
    \end{subfigure}
    \hfill
    \begin{subfigure}[c]{0.25\textwidth}
        \centering
        \includesvg[inkscapelatex=false, width=\textwidth]{mapper_concentric_circles_hdbscan_15_intervals_0.3_overlap}
        \caption{}
        \label{fig:mapper_concentric_circles_hdbscan}
    \end{subfigure}
    \caption{(a) input data set; result of Mapper with (b) \aut; (c) DBSCAN; (d) HDBSCAN}
    \label{fig:mapper_concentric_circles}
\end{figure}

We ran the Mapper implementation of \emph{giotto-tda}~\citep{giotto-tda} on a synthetic two-dimensional data set consisting of noisy samples from two concentric circles (see Figure~\ref{fig:concentric_circles}) with projection onto the $x$-axis as the filter function.
We ran Mapper on the same interval cover with three different choices of clustering algorithms: \aut, DBSCAN, and HDBSCAN. As can be seen in Figure~\ref{fig:mapper_concentric_circles_automato}, using DBSCAN, we see many unwanted edges in the graph. HDBSCAN performs better, giving two cycles with some extra loops.
The output of Mapper with \aut is exactly the Reeb graph of two circles.

\begin{figure}[ht]
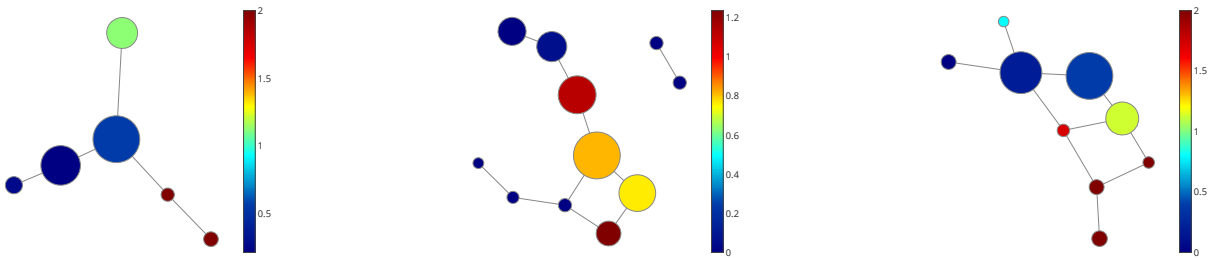

    \centering
    \begin{subfigure}{0.25\textwidth}
        \centering
        \includesvg[inkscapelatex=false, width=\textwidth]{mapper_diabetes_automato_4_intervals_0.5_overlap}
        \label{fig:mapper_diabetes_automato_4_intervals_0.5_overlap}
    \end{subfigure}
    \hfill
    \begin{subfigure}{0.25\textwidth}
        \centering
        \includesvg[inkscapelatex=false, width=\textwidth]{mapper_diabetes_dbscan_4_intervals_0.5_overlap}
        \label{fig:mapper_diabetes_dbscan_4_intervals_0.5_overlap}
    \end{subfigure}
    \hfill
    \begin{subfigure}{0.25\textwidth}
        \centering
        \includesvg[inkscapelatex=false, width=\textwidth]{mapper_diabetes_hdbscan_4_intervals_0.5_overlap}
    \label{fig:mapper_diabetes_hdbscan_4_intervals_0.5_overlap}
    \end{subfigure}
    \caption{Mapper applied to the diabetes data set with \aut (left); DBSCAN (center); HDBSCAN (right). Labels 0, 1 and 2 stand for ``no '', ``chemical'' and ``overt diabetes''.}
    \label{fig:mapper_diabetes}
\end{figure}

We further tested the combination of Mapper with \aut on one of the standard applications of Mapper: the Miller-Reaven diabetes data set, where Mapper can be used detect two strains of diabetes that correspond to ``flares'' in the data set (see \citet[Section 5.1]{singh2007topological} for details).\footnote{The data set is available as part of the ``locfit'' R-package~\citep{locfit_package}.} As can be seen in Figure~\ref{fig:mapper_diabetes}, \aut performs well in this task; the graphs show a central core of vertices corresponding to healthy patients, and two flares corresponding to the two strains of diabetes. We were not able to reproduce this using DBSCAN or HDBSCAN; Figure~\ref{fig:mapper_diabetes} shows the output of Mapper with these algorithms with their respective default parameters.

%%%%%%%%%%%%%%%%%%%%%%%%%%%%%%%%%%%%%%%%%%%%%%%%%%%%%%%%%%%%%%%%%%%%%%%%%%%%%%%%%%%%%%%%%%%%%%%%

\section{Discussion}\label{sec:discussion}

We briefly outline some limitations of \aut.
\aut comes with a choice of default values for its parameters. 

Optimizing the choice of the neighborhood graph and density estimators is an aspect of \aut that we plan to pursue in future work. Moreover, we plan to improve the currently experimental feature for outlier creation in \aut discussed at the end of Section~\ref{subsec:implementation}. Finally, it is natural to ask whether the results from~\citet{mapper_parameter_selection_carriere} on optimal parameter selection in the Mapper algorithm can be adapted to the scenario where Mapper uses \aut as its clustering algorithm.

%%%%%%%%%%%%%%%%%%%%%%%%%%%%%%%%%%%%%%%%%%%%%%%%%%%%%%%%%%%%%%%%%%%%%%%%%%%%%%%%%%%%%%%%%%%%%%%%

\subsubsection*{Acknowledgments}

The first author was supported by the Swiss National Science Foundation (project no. 209413).

%%%%%%%%%%%%%%%%%%%%%%%%%%%%%%%%%%%%%%%%%%%%%%%%%%%%%%%%%%%%%%%%%%%%%%%%%%%%%%%%%%%%%%%%%%%%%%%%

\bibliography{biblio}
\bibliographystyle{tmlr}

%%%%%%%%%%%%%%%%%%%%%%%%%%%%%%%%%%%%%%%%%%%%%%%%%%%%%%%%%%%%%%%%%%%%%%%%%%%%%%%%%%%%%%%%%%%%%%%%

\newpage

\appendix

\section{Appendix}\label{sec:appendix}

\subsection{About the choice of data sets}

\begin{figure}[ht]
    \centering
    \includegraphics[width=0.4\textwidth]{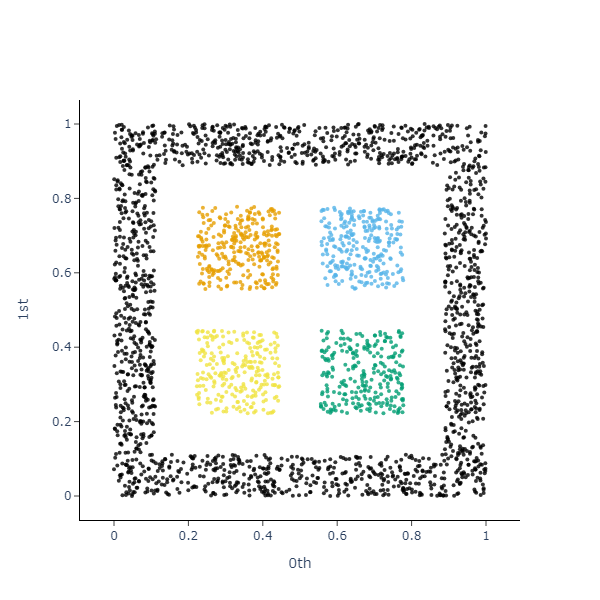}
    \includegraphics[width=0.4\textwidth]{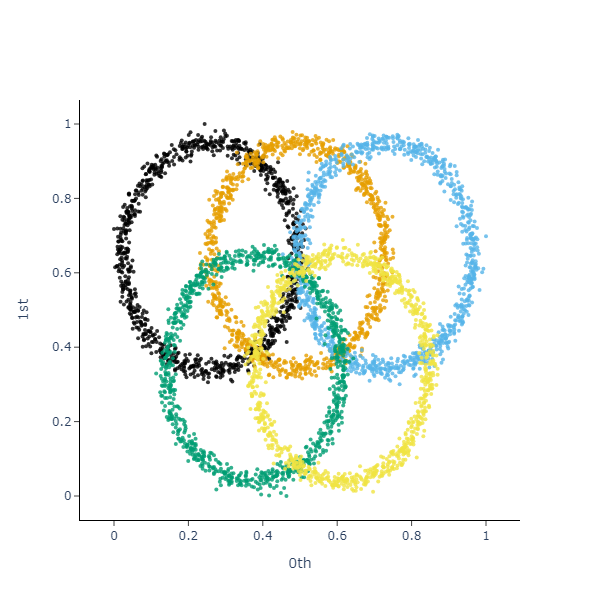}
    \caption{The data sets named \texttt{windows} (left) and \texttt{olympic} (right) from the \texttt{wut}-battery.}
    \label{fig:wut_and_olympic}
\end{figure}

As explained in Section~\ref{subsec:choice_of_data_sets_for_comparison}, we chose to include the data set named \texttt{windows} from the battery named \texttt{wut}, but not the data set named \texttt{olympic} from the same battery. Those are illustrated in Figure~\ref{fig:wut_and_olympic}. In that figure, the data points are colored according to the ground truth clustering.

\subsection{Benchmarking results}\label{appendix_subsec:benchmarking_results}

In this subsection we report the Fowlkes-Mallows scores coming from comparing \aut to the other clustering algorithms, as explained in Section~\ref{sec:experiments}. For those benchmarking data sets that come with more than one ground truth, we report the scores for each of those, and different ground truths are indicated by the last digit in the data set name. Moreover, each table is sorted according to increasing difference in clustering performance of \aut and the respective clustering algorithm that \aut is being compared against. As is customary, we indicate the score stemming from the best performing clustering algorithm in bold.
Finally, each of the table is accompanied by a graph similar to the one depicted in Figure~\ref{fig:aut_vs_dbscan_mainbody}. Note that, in particular, that those figures indicate only the score corresponding to the ground truth on which the respective clustering algorithm performs best on.

\begin{longtable}{lrrrr}
\caption{Fowlkes-Mallows scores of AuToMATo vs. DBSCAN}
\label{table:aut_vs_dbscan} \\
\toprule
Dataset & automato\_mean & dbscan\_max & dbscan\_min \\
\midrule
\endfirsthead
\caption[]{Fowlkes-Mallows scores of AuToMATo vs. DBSCAN} \\
\toprule
Dataset & automato\_mean & dbscan\_max & dbscan\_min \\
\midrule
\endhead
\midrule
\multicolumn{5}{r}{Continued on next page} \\
\midrule
\endfoot
\bottomrule
\endlastfoot
sipu\_r15\_2 & 0.4867±0.0000 & \textbf{1.0000} & 0.5607 \\
wut\_trajectories\_0 & 0.5038±0.0107 & \textbf{1.0000} & 0.4999 \\
wut\_x3\_0 & 0.5153±0.0000 & \textbf{0.9398} & 0.5149 \\
wut\_x2\_0 & 0.5846±0.0000 & \textbf{0.9483} & 0.5779 \\
sipu\_r15\_1 & 0.5436±0.0000 & \textbf{0.8954} & 0.5021 \\
fcps\_tetra\_0 & 0.6261±0.0000 & \textbf{0.9403} & 0.0000 \\
sipu\_pathbased\_0 & 0.6517±0.0000 & \textbf{0.9569} & 0.5769 \\
sipu\_spiral\_0 & 0.7028±0.0000 & \textbf{1.0000} & 0.5756 \\
wut\_isolation\_0 & 0.7256±0.0113 & \textbf{1.0000} & 0.5773 \\
sipu\_pathbased\_1 & 0.7322±0.0000 & \textbf{0.9620} & 0.5170 \\
sipu\_jain\_0 & 0.7837±0.0000 & \textbf{0.9880} & 0.7837 \\
graves\_dense\_0 & 0.8377±0.1396 & \textbf{0.9970} & 0.7053 \\
sipu\_compound\_0 & 0.8616±0.0000 & \textbf{1.0000} & 0.4972 \\
fcps\_atom\_0 & 0.8694±0.0000 & \textbf{1.0000} & 0.7067 \\
wut\_circles\_0 & 0.8857±0.0000 & \textbf{1.0000} & 0.4998 \\
fcps\_chainlink\_0 & 0.8896±0.0000 & \textbf{1.0000} & 0.7068 \\
other\_iris\_0 & 0.7715±0.0000 & \textbf{0.8721} & 0.0000 \\
wut\_mk4\_0 & 0.9072±0.0234 & \textbf{1.0000} & 0.5770 \\
wut\_mk2\_0 & 0.6356±0.0000 & \textbf{0.7068} & 0.5778 \\
sipu\_compound\_4 & 0.9442±0.0000 & \textbf{1.0000} & 0.5523 \\
wut\_x3\_1 & 0.6546±0.0000 & \textbf{0.7042} & 0.6546 \\
graves\_zigzag\_1 & 0.6720±0.0000 & \textbf{0.7149} & 0.4446 \\
other\_iris5\_0 & 0.6712±0.0000 & \textbf{0.7046} & 0.0000 \\
wut\_smile\_1 & 0.9701±0.0000 & \textbf{1.0000} & 0.5825 \\
wut\_x1\_0 & 0.9741±0.0818 & \textbf{1.0000} & 0.5846 \\
fcps\_target\_0 & 0.9850±0.0000 & \textbf{1.0000} & 0.6963 \\
wut\_smile\_0 & 0.9681±0.0000 & \textbf{0.9753} & 0.5471 \\
wut\_mk3\_0 & 0.7720±0.0000 & \textbf{0.7774} & 0.5764 \\
sipu\_compound\_1 & 0.9786±0.0000 & \textbf{0.9825} & 0.5715 \\
sipu\_flame\_0 & 0.7320±0.0000 & \textbf{0.7341} & 0.5918 \\
sipu\_unbalance\_0 & 0.9986±0.0008 & \textbf{1.0000} & 0.5339 \\
fcps\_twodiamonds\_0 & \textbf{0.7067±0.0000} & \textbf{0.7067} & \textbf{0.7067} \\
sipu\_aggregation\_0 & \textbf{0.8652±0.0000} & \textbf{0.8652} & 0.4653 \\
wut\_stripes\_0 & \textbf{1.0000±0.0000} & \textbf{1.0000} & 0.7070 \\
wut\_trapped\_lovers\_0 & \textbf{1.0000±0.0000} & \textbf{1.0000} & 0.6632 \\
wut\_windows\_0 & \textbf{1.0000±0.0000} & \textbf{1.0000} & 0.6753 \\
fcps\_hepta\_0 & \textbf{1.0000±0.0000} & \textbf{1.0000} & 0.3727 \\
fcps\_lsun\_0 & \textbf{1.0000±0.0000} & \textbf{1.0000} & 0.6111 \\
graves\_line\_0 & \textbf{1.0000±0.0000} & \textbf{1.0000} & 0.8238 \\
graves\_ring\_0 & \textbf{1.0000±0.0000} & \textbf{1.0000} & 0.7068 \\
graves\_ring\_outliers\_0 & \textbf{1.0000±0.0000} & \textbf{1.0000} & 0.6863 \\
graves\_zigzag\_0 & \textbf{1.0000±0.0000} & \textbf{1.0000} & 0.5328 \\
other\_square\_0 & \textbf{1.0000±0.0000} & \textbf{1.0000} & 0.7068 \\
wut\_mk1\_0 & \textbf{0.9866±0.0000} & 0.9651 & 0.5754 \\
wut\_twosplashes\_0 & \textbf{1.0000±0.0000} & 0.9649 & 0.7062 \\
fcps\_wingnut\_0 & \textbf{0.9805±0.0000} & 0.8784 & 0.7068 \\
graves\_parabolic\_1 & \textbf{0.6916±0.0000} & 0.5000 & 0.4999 \\
wut\_labirynth\_0 & \textbf{0.7884±0.0000} & 0.5221 & 0.5221 \\
graves\_parabolic\_0 & \textbf{0.9802±0.0000} & 0.7068 & 0.7068 \\
sipu\_d31\_0 & \textbf{0.6001±0.0085} & 0.1846 & 0.1787 \\
sipu\_a1\_0 & \textbf{0.7499±0.0000} & 0.3269 & 0.2229 \\
sipu\_r15\_0 & \textbf{0.9258±0.0000} & 0.4551 & 0.2552 \\
sipu\_s1\_0 & \textbf{0.9888±0.0000} & 0.4890 & 0.2581 \\
sipu\_a2\_0 & \textbf{0.7555±0.0000} & 0.1685 & 0.1685 \\
sipu\_a3\_0 & \textbf{0.7434±0.0000} & 0.1410 & 0.1410 \\
sipu\_s2\_0 & \textbf{0.9405±0.0000} & 0.2581 & 0.2581 \\
\end{longtable}

\begin{figure}[ht]
    \centering
    \includesvg[inkscapelatex=false, width=\textwidth]{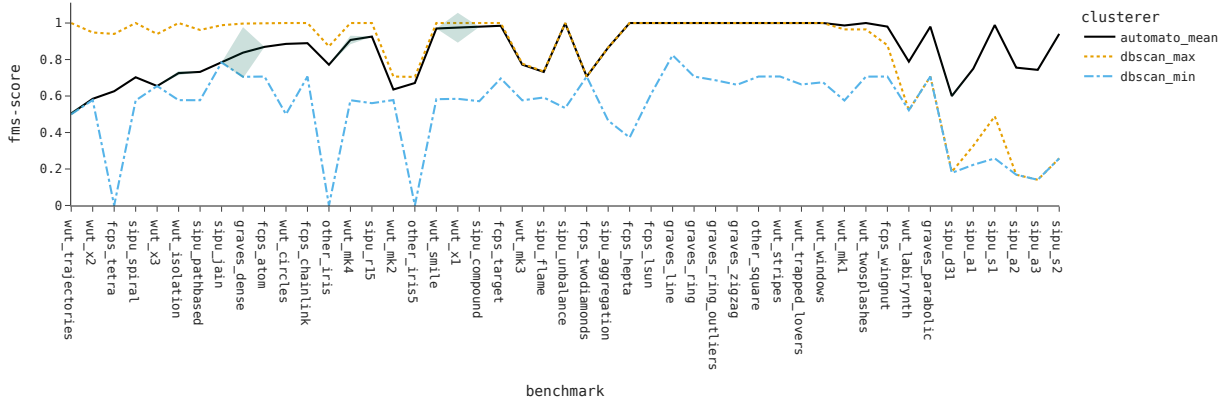}
    \caption{Comparison of \aut and DBSCAN.}
    \label{fig:aut_vs_dbscan}
\end{figure}

\begin{longtable}{lrrrr}
\caption{Fowlkes-Mallows scores of AuToMATo vs. hierarchical clustering with average linkage}
\label{table:aut_vs_linkage_average} \\
\toprule
Dataset & automato\_mean & linkage\_average\_max & linkage\_average\_min \\
\midrule
\endfirsthead
\caption[]{Fowlkes-Mallows scores of AuToMATo vs. hierarchical clustering with average linkage} \\
\toprule
Dataset & automato\_mean & linkage\_average\_max & linkage\_average\_min \\
\midrule
\endhead
\midrule
\multicolumn{5}{r}{Continued on next page} \\
\midrule
\endfoot
\bottomrule
\endlastfoot
sipu\_r15\_2 & 0.4867±0.0000 & \textbf{1.0000} & 0.3971 \\
wut\_trajectories\_0 & 0.5038±0.0107 & \textbf{1.0000} & 0.3115 \\
fcps\_tetra\_0 & 0.6261±0.0000 & \textbf{1.0000} & 0.0651 \\
sipu\_r15\_1 & 0.5436±0.0000 & \textbf{0.8954} & 0.4435 \\
sipu\_d31\_0 & 0.6001±0.0085 & \textbf{0.9322} & 0.1787 \\
wut\_x3\_0 & 0.5153±0.0000 & \textbf{0.8389} & 0.3343 \\
wut\_x3\_1 & 0.6546±0.0000 & \textbf{0.9747} & 0.2693 \\
fcps\_twodiamonds\_0 & 0.7067±0.0000 & \textbf{0.9925} & 0.1287 \\
sipu\_a2\_0 & 0.7555±0.0000 & \textbf{0.9432} & 0.1685 \\
sipu\_a1\_0 & 0.7499±0.0000 & \textbf{0.9268} & 0.2229 \\
sipu\_a3\_0 & 0.7434±0.0000 & \textbf{0.8825} & 0.1410 \\
sipu\_aggregation\_0 & 0.8652±0.0000 & \textbf{0.9932} & 0.1785 \\
sipu\_pathbased\_1 & 0.7322±0.0000 & \textbf{0.8564} & 0.2058 \\
graves\_dense\_0 & 0.8377±0.1396 & \textbf{0.9604} & 0.6633 \\
sipu\_pathbased\_0 & 0.6517±0.0000 & \textbf{0.7704} & 0.1848 \\
wut\_x2\_0 & 0.5846±0.0000 & \textbf{0.7001} & 0.2782 \\
wut\_circles\_0 & 0.8857±0.0000 & \textbf{1.0000} & 0.2369 \\
wut\_mk3\_0 & 0.7720±0.0000 & \textbf{0.8771} & 0.0876 \\
wut\_mk2\_0 & 0.6356±0.0000 & \textbf{0.7068} & 0.1421 \\
sipu\_r15\_0 & 0.9258±0.0000 & \textbf{0.9900} & 0.2552 \\
graves\_zigzag\_1 & 0.6720±0.0000 & \textbf{0.7202} & 0.4380 \\
other\_iris\_0 & 0.7715±0.0000 & \textbf{0.8080} & 0.0791 \\
other\_iris5\_0 & 0.6712±0.0000 & \textbf{0.7042} & 0.0451 \\
wut\_x1\_0 & 0.9741±0.0818 & \textbf{1.0000} & 0.3070 \\
sipu\_jain\_0 & 0.7837±0.0000 & \textbf{0.7904} & 0.1736 \\
wut\_mk1\_0 & 0.9866±0.0000 & \textbf{0.9933} & 0.2037 \\
sipu\_unbalance\_0 & 0.9986±0.0008 & \textbf{0.9995} & 0.5339 \\
fcps\_hepta\_0 & \textbf{1.0000±0.0000} & \textbf{1.0000} & 0.3727 \\
sipu\_flame\_0 & \textbf{0.7320±0.0000} & \textbf{0.7320} & 0.0913 \\
sipu\_s1\_0 & \textbf{0.9888±0.0000} & 0.9821 & 0.2581 \\
sipu\_compound\_0 & \textbf{0.8616±0.0000} & 0.8431 & 0.2207 \\
sipu\_compound\_4 & \textbf{0.9442±0.0000} & 0.9224 & 0.1985 \\
sipu\_compound\_1 & \textbf{0.9786±0.0000} & 0.9546 & 0.1922 \\
sipu\_s2\_0 & \textbf{0.9405±0.0000} & 0.9097 & 0.2581 \\
wut\_smile\_1 & \textbf{0.9701±0.0000} & 0.8726 & 0.4041 \\
fcps\_atom\_0 & \textbf{0.8694±0.0000} & 0.7491 & 0.2555 \\
graves\_parabolic\_1 & \textbf{0.6916±0.0000} & 0.5708 & 0.2135 \\
sipu\_spiral\_0 & \textbf{0.7028±0.0000} & 0.5756 & 0.1919 \\
wut\_mk4\_0 & \textbf{0.9072±0.0234} & 0.7714 & 0.2071 \\
wut\_smile\_0 & \textbf{0.9681±0.0000} & 0.8221 & 0.4303 \\
wut\_isolation\_0 & \textbf{0.7256±0.0113} & 0.5773 & 0.1651 \\
graves\_line\_0 & \textbf{1.0000±0.0000} & 0.8238 & 0.3047 \\
fcps\_chainlink\_0 & \textbf{0.8896±0.0000} & 0.7068 & 0.1456 \\
fcps\_target\_0 & \textbf{0.9850±0.0000} & 0.7986 & 0.3285 \\
fcps\_wingnut\_0 & \textbf{0.9805±0.0000} & 0.7739 & 0.1101 \\
fcps\_lsun\_0 & \textbf{1.0000±0.0000} & 0.7896 & 0.1735 \\
graves\_ring\_0 & \textbf{1.0000±0.0000} & 0.7780 & 0.2638 \\
graves\_parabolic\_0 & \textbf{0.9802±0.0000} & 0.7580 & 0.1598 \\
graves\_ring\_outliers\_0 & \textbf{1.0000±0.0000} & 0.7767 & 0.2801 \\
other\_square\_0 & \textbf{1.0000±0.0000} & 0.7413 & 0.1746 \\
wut\_labirynth\_0 & \textbf{0.7884±0.0000} & 0.5221 & 0.2306 \\
wut\_stripes\_0 & \textbf{1.0000±0.0000} & 0.7070 & 0.1082 \\
wut\_twosplashes\_0 & \textbf{1.0000±0.0000} & 0.7062 & 0.4837 \\
wut\_windows\_0 & \textbf{1.0000±0.0000} & 0.6753 & 0.1194 \\
wut\_trapped\_lovers\_0 & \textbf{1.0000±0.0000} & 0.6632 & 0.1077 \\
graves\_zigzag\_0 & \textbf{1.0000±0.0000} & 0.6616 & 0.3008 \\
\end{longtable}

\begin{figure}[ht]
    \centering
    \includesvg[inkscapelatex=false, width=\textwidth]{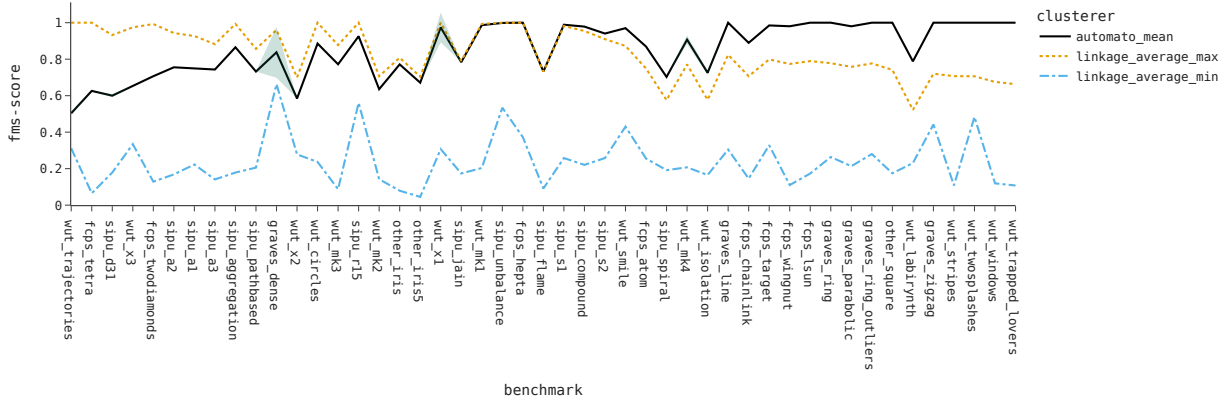}
    \caption{Comparison of \aut and agglomerative clustering with average linkage.}
    \label{fig:aut_vs_linkage_average}
\end{figure}

\begin{longtable}{lrrr}
\caption{Fowlkes-Mallows scores of AuToMATo vs. HDBSCAN}
\label{table:aut_vs_hdbscan} \\
\toprule
Dataset & automato\_mean & automato\_std & hdbscan \\
\midrule
\endfirsthead
\caption[]{Fowlkes-Mallows scores of AuToMATo vs. HDBSCAN} \\
\toprule
Dataset & automato\_mean & automato\_std & hdbscan \\
\midrule
\endhead
\midrule
\multicolumn{4}{r}{Continued on next page} \\
\midrule
\endfoot
\bottomrule
\endlastfoot
wut\_trajectories\_0 & 0.5038±0.0107 & 0.0107 & \textbf{1.0000} \\
wut\_x3\_0 & 0.5153±0.0000 & 0.0000 & \textbf{0.8959} \\
wut\_x2\_0 & 0.5846±0.0000 & 0.0000 & \textbf{0.9344} \\
sipu\_spiral\_0 & 0.7028±0.0000 & 0.0000 & \textbf{0.9815} \\
sipu\_flame\_0 & 0.7320±0.0000 & 0.0000 & \textbf{0.9900} \\
sipu\_d31\_0 & 0.6001±0.0085 & 0.0085 & \textbf{0.8231} \\
sipu\_jain\_0 & 0.7837±0.0000 & 0.0000 & \textbf{0.9779} \\
fcps\_tetra\_0 & 0.6261±0.0000 & 0.0000 & \textbf{0.8157} \\
graves\_dense\_0 & 0.8377±0.1396 & 0.1396 & \textbf{0.9894} \\
sipu\_pathbased\_1 & 0.7322±0.0000 & 0.0000 & \textbf{0.8634} \\
fcps\_atom\_0 & 0.8694±0.0000 & 0.0000 & \textbf{1.0000} \\
sipu\_pathbased\_0 & 0.6517±0.0000 & 0.0000 & \textbf{0.7815} \\
fcps\_chainlink\_0 & 0.8896±0.0000 & 0.0000 & \textbf{1.0000} \\
sipu\_r15\_0 & 0.9258±0.0000 & 0.0000 & \textbf{0.9932} \\
sipu\_a1\_0 & 0.7499±0.0000 & 0.0000 & \textbf{0.8081} \\
wut\_x3\_1 & 0.6546±0.0000 & 0.0000 & \textbf{0.6972} \\
other\_iris5\_0 & 0.6712±0.0000 & 0.0000 & \textbf{0.7042} \\
wut\_x1\_0 & 0.9741±0.0818 & 0.0818 & \textbf{1.0000} \\
sipu\_compound\_1 & 0.9786±0.0000 & 0.0000 & \textbf{1.0000} \\
sipu\_compound\_4 & 0.9442±0.0000 & 0.0000 & \textbf{0.9656} \\
fcps\_target\_0 & 0.9850±0.0000 & 0.0000 & \textbf{1.0000} \\
sipu\_compound\_0 & 0.8616±0.0000 & 0.0000 & \textbf{0.8751} \\
sipu\_unbalance\_0 & 0.9986±0.0008 & 0.0008 & \textbf{1.0000} \\
graves\_zigzag\_1 & \textbf{0.6720±0.0000} & 0.0000 & \textbf{0.6720} \\
sipu\_aggregation\_0 & \textbf{0.8652±0.0000} & 0.0000 & \textbf{0.8652} \\
wut\_stripes\_0 & \textbf{1.0000±0.0000} & 0.0000 & \textbf{1.0000} \\
wut\_trapped\_lovers\_0 & \textbf{1.0000±0.0000} & 0.0000 & \textbf{1.0000} \\
wut\_windows\_0 & \textbf{1.0000±0.0000} & 0.0000 & \textbf{1.0000} \\
fcps\_hepta\_0 & \textbf{1.0000±0.0000} & 0.0000 & \textbf{1.0000} \\
fcps\_lsun\_0 & \textbf{1.0000±0.0000} & 0.0000 & \textbf{1.0000} \\
graves\_line\_0 & \textbf{1.0000±0.0000} & 0.0000 & \textbf{1.0000} \\
graves\_ring\_0 & \textbf{1.0000±0.0000} & 0.0000 & \textbf{1.0000} \\
graves\_ring\_outliers\_0 & \textbf{1.0000±0.0000} & 0.0000 & \textbf{1.0000} \\
graves\_zigzag\_0 & \textbf{1.0000±0.0000} & 0.0000 & \textbf{1.0000} \\
other\_square\_0 & \textbf{1.0000±0.0000} & 0.0000 & \textbf{1.0000} \\
other\_iris\_0 & \textbf{0.7715±0.0000} & 0.0000 & \textbf{0.7715} \\
wut\_mk3\_0 & \textbf{0.7720±0.0000} & 0.0000 & 0.7719 \\
wut\_mk1\_0 & \textbf{0.9866±0.0000} & 0.0000 & 0.9863 \\
sipu\_a3\_0 & \textbf{0.7434±0.0000} & 0.0000 & 0.7415 \\
sipu\_a2\_0 & \textbf{0.7555±0.0000} & 0.0000 & 0.7502 \\
sipu\_r15\_2 & \textbf{0.4867±0.0000} & 0.0000 & 0.4671 \\
sipu\_r15\_1 & \textbf{0.5436±0.0000} & 0.0000 & 0.5212 \\
wut\_isolation\_0 & \textbf{0.7256±0.0113} & 0.0113 & 0.6377 \\
fcps\_wingnut\_0 & \textbf{0.9805±0.0000} & 0.0000 & 0.8725 \\
sipu\_s1\_0 & \textbf{0.9888±0.0000} & 0.0000 & 0.8717 \\
sipu\_s2\_0 & \textbf{0.9405±0.0000} & 0.0000 & 0.7410 \\
wut\_mk4\_0 & \textbf{0.9072±0.0234} & 0.0234 & 0.6459 \\
wut\_labirynth\_0 & \textbf{0.7884±0.0000} & 0.0000 & 0.5134 \\
graves\_parabolic\_1 & \textbf{0.6916±0.0000} & 0.0000 & 0.3616 \\
fcps\_twodiamonds\_0 & \textbf{0.7067±0.0000} & 0.0000 & 0.2886 \\
wut\_mk2\_0 & \textbf{0.6356±0.0000} & 0.0000 & 0.1574 \\
wut\_smile\_0 & \textbf{0.9681±0.0000} & 0.0000 & 0.4000 \\
wut\_smile\_1 & \textbf{0.9701±0.0000} & 0.0000 & 0.3714 \\
graves\_parabolic\_0 & \textbf{0.9802±0.0000} & 0.0000 & 0.3526 \\
wut\_twosplashes\_0 & \textbf{1.0000±0.0000} & 0.0000 & 0.3074 \\
wut\_circles\_0 & \textbf{0.8857±0.0000} & 0.0000 & 0.1204 \\
\end{longtable}

\begin{figure}[ht]
    \centering
    \includesvg[inkscapelatex=false, width=\textwidth]{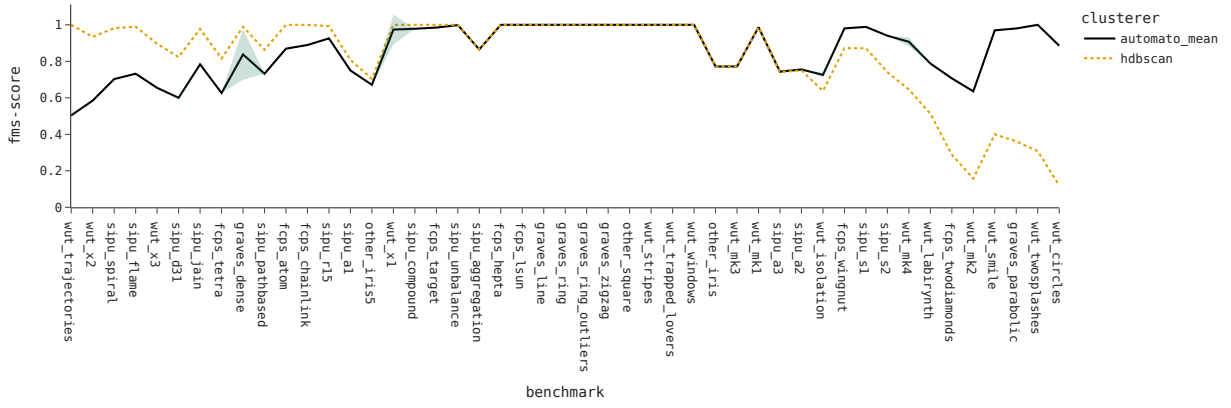}
    \caption{Comparison of \aut and HDBSCAN.}
    \label{fig:aut_vs_hdbscan}
\end{figure}

\begin{longtable}{lrrrr}
\caption{Fowlkes-Mallows scores of AuToMATo vs. hierarchical clustering with single linkage}
\label{table:aut_vs_linkage_single} \\
\toprule
Dataset & automato\_mean & linkage\_single\_max & linkage\_single\_min \\
\midrule
\endfirsthead
\caption[]{Fowlkes-Mallows scores of AuToMATo vs. hierarchical clustering with single linkage} \\
\toprule
Dataset & automato\_mean & linkage\_single\_max & linkage\_single\_min \\
\midrule
\endhead
\midrule
\multicolumn{5}{r}{Continued on next page} \\
\midrule
\endfoot
\bottomrule
\endlastfoot
sipu\_r15\_2 & 0.4867±0.0000 & \textbf{1.0000} & 0.5607 \\
wut\_trajectories\_0 & 0.5038±0.0107 & \textbf{1.0000} & 0.4999 \\
sipu\_r15\_1 & 0.5436±0.0000 & \textbf{0.8954} & 0.5021 \\
fcps\_tetra\_0 & 0.6261±0.0000 & \textbf{0.9296} & 0.0829 \\
sipu\_spiral\_0 & 0.7028±0.0000 & \textbf{1.0000} & 0.5756 \\
wut\_isolation\_0 & 0.7256±0.0113 & \textbf{1.0000} & 0.5773 \\
wut\_x3\_0 & 0.5153±0.0000 & \textbf{0.7347} & 0.4951 \\
sipu\_jain\_0 & 0.7837±0.0000 & \textbf{0.9510} & 0.7837 \\
fcps\_atom\_0 & 0.8694±0.0000 & \textbf{1.0000} & 0.7067 \\
wut\_circles\_0 & 0.8857±0.0000 & \textbf{1.0000} & 0.4998 \\
fcps\_chainlink\_0 & 0.8896±0.0000 & \textbf{1.0000} & 0.7068 \\
wut\_mk4\_0 & 0.9072±0.0234 & \textbf{1.0000} & 0.5770 \\
sipu\_compound\_0 & 0.8616±0.0000 & \textbf{0.9454} & 0.4972 \\
sipu\_pathbased\_0 & 0.6517±0.0000 & \textbf{0.7337} & 0.5769 \\
sipu\_pathbased\_1 & 0.7322±0.0000 & \textbf{0.8091} & 0.5170 \\
graves\_dense\_0 & 0.8377±0.1396 & \textbf{0.9096} & 0.6882 \\
wut\_mk2\_0 & 0.6356±0.0000 & \textbf{0.7068} & 0.6007 \\
graves\_zigzag\_1 & 0.6720±0.0000 & \textbf{0.7344} & 0.4446 \\
wut\_x2\_0 & 0.5846±0.0000 & \textbf{0.6437} & 0.5105 \\
other\_iris5\_0 & 0.6712±0.0000 & \textbf{0.7042} & 0.0451 \\
wut\_smile\_1 & 0.9701±0.0000 & \textbf{1.0000} & 0.5825 \\
wut\_x1\_0 & 0.9741±0.0818 & \textbf{0.9920} & 0.5846 \\
fcps\_target\_0 & 0.9850±0.0000 & \textbf{1.0000} & 0.6963 \\
wut\_smile\_0 & 0.9681±0.0000 & \textbf{0.9748} & 0.5471 \\
sipu\_unbalance\_0 & 0.9986±0.0008 & \textbf{1.0000} & 0.5339 \\
fcps\_twodiamonds\_0 & \textbf{0.7067±0.0000} & \textbf{0.7067} & \textbf{0.7067} \\
wut\_x3\_1 & \textbf{0.6546±0.0000} & \textbf{0.6546} & 0.6140 \\
sipu\_aggregation\_0 & \textbf{0.8652±0.0000} & \textbf{0.8652} & 0.4653 \\
wut\_stripes\_0 & \textbf{1.0000±0.0000} & \textbf{1.0000} & 0.7070 \\
wut\_trapped\_lovers\_0 & \textbf{1.0000±0.0000} & \textbf{1.0000} & 0.6632 \\
wut\_windows\_0 & \textbf{1.0000±0.0000} & \textbf{1.0000} & 0.6753 \\
fcps\_hepta\_0 & \textbf{1.0000±0.0000} & \textbf{1.0000} & 0.3727 \\
graves\_line\_0 & \textbf{1.0000±0.0000} & \textbf{1.0000} & 0.8238 \\
graves\_ring\_0 & \textbf{1.0000±0.0000} & \textbf{1.0000} & 0.7068 \\
graves\_ring\_outliers\_0 & \textbf{1.0000±0.0000} & \textbf{1.0000} & 0.6863 \\
graves\_zigzag\_0 & \textbf{1.0000±0.0000} & \textbf{1.0000} & 0.5381 \\
sipu\_flame\_0 & \textbf{0.7320±0.0000} & \textbf{0.7320} & 0.4598 \\
other\_iris\_0 & \textbf{0.7715±0.0000} & \textbf{0.7715} & 0.1223 \\
other\_square\_0 & \textbf{1.0000±0.0000} & 0.9990 & 0.7068 \\
fcps\_lsun\_0 & \textbf{1.0000±0.0000} & 0.9983 & 0.6111 \\
wut\_twosplashes\_0 & \textbf{1.0000±0.0000} & 0.9850 & 0.7062 \\
sipu\_compound\_1 & \textbf{0.9786±0.0000} & 0.9180 & 0.5715 \\
sipu\_compound\_4 & \textbf{0.9442±0.0000} & 0.8824 & 0.5523 \\
wut\_mk1\_0 & \textbf{0.9866±0.0000} & 0.8866 & 0.5754 \\
fcps\_wingnut\_0 & \textbf{0.9805±0.0000} & 0.8087 & 0.7068 \\
graves\_parabolic\_1 & \textbf{0.6916±0.0000} & 0.5000 & 0.4979 \\
wut\_mk3\_0 & \textbf{0.7720±0.0000} & 0.5764 & 0.5314 \\
wut\_labirynth\_0 & \textbf{0.7884±0.0000} & 0.5221 & 0.5221 \\
graves\_parabolic\_0 & \textbf{0.9802±0.0000} & 0.7068 & 0.7040 \\
sipu\_d31\_0 & \textbf{0.6001±0.0085} & 0.1846 & 0.1787 \\
sipu\_a1\_0 & \textbf{0.7499±0.0000} & 0.3269 & 0.2229 \\
sipu\_r15\_0 & \textbf{0.9258±0.0000} & 0.4551 & 0.2552 \\
sipu\_a2\_0 & \textbf{0.7555±0.0000} & 0.1685 & 0.1685 \\
sipu\_a3\_0 & \textbf{0.7434±0.0000} & 0.1410 & 0.1410 \\
sipu\_s1\_0 & \textbf{0.9888±0.0000} & 0.3695 & 0.2581 \\
sipu\_s2\_0 & \textbf{0.9405±0.0000} & 0.2581 & 0.2579 \\
\end{longtable}

\begin{figure}[ht]
    \centering
    \includesvg[inkscapelatex=false, width=\textwidth]{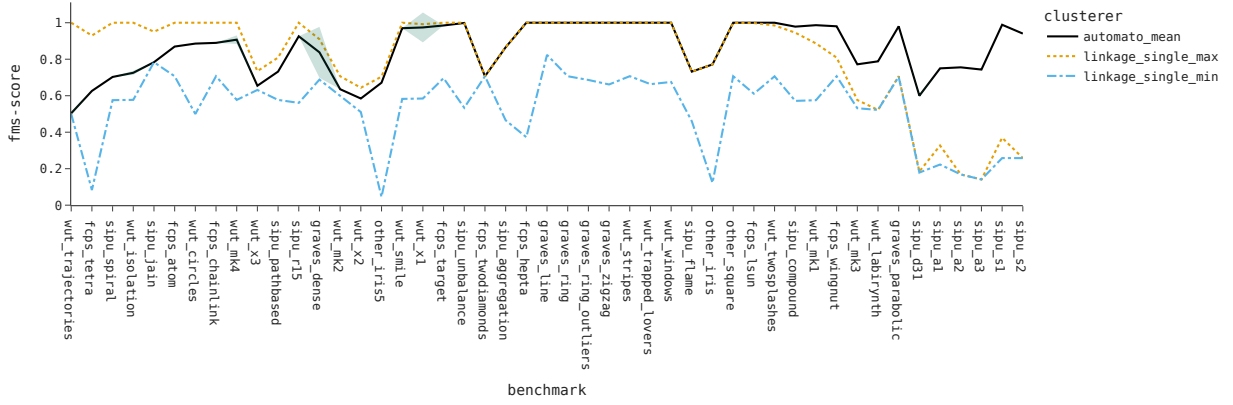}
    \caption{Comparison of \aut and agglomerative clustering with single linkage.}
    \label{fig:aut_vs_linkage_single}
\end{figure}

\begin{longtable}{lrrr}
\caption{Fowlkes-Mallows scores of AuToMATo vs. TTK clustering algorithm}
\label{table:aut_vs_ttk} \\
\toprule
Dataset & automato\_mean & automato\_std & ttk \\
\midrule
\endfirsthead
\caption[]{Fowlkes-Mallows scores of AuToMATo vs. TTK clustering algorithm} \\
\toprule
Dataset & automato\_mean & automato\_std & ttk \\
\midrule
\endhead
\midrule
\multicolumn{4}{r}{Continued on next page} \\
\midrule
\endfoot
\bottomrule
\endlastfoot
wut\_trajectories\_0 & 0.5038±0.0107 & 0.0107 & \textbf{0.8682} \\
fcps\_tetra\_0 & 0.6261±0.0000 & 0.0000 & \textbf{0.9043} \\
wut\_x3\_0 & 0.5153±0.0000 & 0.0000 & \textbf{0.7818} \\
wut\_isolation\_0 & 0.7256±0.0113 & 0.0113 & \textbf{0.9416} \\
sipu\_a1\_0 & 0.7499±0.0000 & 0.0000 & \textbf{0.9143} \\
wut\_x2\_0 & 0.5846±0.0000 & 0.0000 & \textbf{0.7283} \\
sipu\_flame\_0 & 0.7320±0.0000 & 0.0000 & \textbf{0.8562} \\
sipu\_aggregation\_0 & 0.8652±0.0000 & 0.0000 & \textbf{0.9692} \\
graves\_zigzag\_1 & 0.6720±0.0000 & 0.0000 & \textbf{0.7698} \\
other\_iris\_0 & 0.7715±0.0000 & 0.0000 & \textbf{0.8639} \\
sipu\_jain\_0 & 0.7837±0.0000 & 0.0000 & \textbf{0.8182} \\
graves\_dense\_0 & 0.8377±0.1396 & 0.1396 & \textbf{0.8615} \\
sipu\_r15\_0 & 0.9258±0.0000 & 0.0000 & \textbf{0.9374} \\
wut\_mk3\_0 & 0.7720±0.0000 & 0.0000 & \textbf{0.7755} \\
wut\_labirynth\_0 & \textbf{0.7884±0.0000} & 0.0000 & \textbf{0.7884} \\
fcps\_chainlink\_0 & \textbf{0.8896±0.0000} & 0.0000 & \textbf{0.8896} \\
wut\_smile\_0 & \textbf{0.9681±0.0000} & 0.0000 & \textbf{0.9681} \\
wut\_stripes\_0 & \textbf{1.0000±0.0000} & 0.0000 & \textbf{1.0000} \\
graves\_ring\_outliers\_0 & \textbf{1.0000±0.0000} & 0.0000 & \textbf{1.0000} \\
wut\_smile\_1 & \textbf{0.9701±0.0000} & 0.0000 & \textbf{0.9701} \\
sipu\_unbalance\_0 & \textbf{0.9986±0.0008} & 0.0008 & 0.9951 \\
sipu\_s1\_0 & \textbf{0.9888±0.0000} & 0.0000 & 0.9843 \\
sipu\_s2\_0 & \textbf{0.9405±0.0000} & 0.0000 & 0.9311 \\
other\_iris5\_0 & \textbf{0.6712±0.0000} & 0.0000 & 0.6612 \\
fcps\_atom\_0 & \textbf{0.8694±0.0000} & 0.0000 & 0.8472 \\
wut\_x3\_1 & \textbf{0.6546±0.0000} & 0.0000 & 0.6312 \\
sipu\_d31\_0 & \textbf{0.6001±0.0085} & 0.0085 & 0.5667 \\
fcps\_hepta\_0 & \textbf{1.0000±0.0000} & 0.0000 & 0.9594 \\
graves\_parabolic\_1 & \textbf{0.6916±0.0000} & 0.0000 & 0.6473 \\
sipu\_r15\_2 & \textbf{0.4867±0.0000} & 0.0000 & 0.4322 \\
sipu\_pathbased\_0 & \textbf{0.6517±0.0000} & 0.0000 & 0.5947 \\
sipu\_spiral\_0 & \textbf{0.7028±0.0000} & 0.0000 & 0.6422 \\
sipu\_r15\_1 & \textbf{0.5436±0.0000} & 0.0000 & 0.4827 \\
sipu\_pathbased\_1 & \textbf{0.7322±0.0000} & 0.0000 & 0.6668 \\
fcps\_target\_0 & \textbf{0.9850±0.0000} & 0.0000 & 0.9185 \\
wut\_x1\_0 & \textbf{0.9741±0.0818} & 0.0818 & 0.8960 \\
wut\_twosplashes\_0 & \textbf{1.0000±0.0000} & 0.0000 & 0.9140 \\
wut\_mk4\_0 & \textbf{0.9072±0.0234} & 0.0234 & 0.8050 \\
wut\_mk2\_0 & \textbf{0.6356±0.0000} & 0.0000 & 0.5302 \\
graves\_parabolic\_0 & \textbf{0.9802±0.0000} & 0.0000 & 0.8653 \\
sipu\_compound\_4 & \textbf{0.9442±0.0000} & 0.0000 & 0.8145 \\
fcps\_wingnut\_0 & \textbf{0.9805±0.0000} & 0.0000 & 0.8497 \\
wut\_circles\_0 & \textbf{0.8857±0.0000} & 0.0000 & 0.7543 \\
wut\_mk1\_0 & \textbf{0.9866±0.0000} & 0.0000 & 0.8148 \\
fcps\_twodiamonds\_0 & \textbf{0.7067±0.0000} & 0.0000 & 0.5251 \\
sipu\_compound\_0 & \textbf{0.8616±0.0000} & 0.0000 & 0.6728 \\
sipu\_compound\_1 & \textbf{0.9786±0.0000} & 0.0000 & 0.7892 \\
graves\_line\_0 & \textbf{1.0000±0.0000} & 0.0000 & 0.7917 \\
fcps\_lsun\_0 & \textbf{1.0000±0.0000} & 0.0000 & 0.7897 \\
wut\_trapped\_lovers\_0 & \textbf{1.0000±0.0000} & 0.0000 & 0.7859 \\
other\_square\_0 & \textbf{1.0000±0.0000} & 0.0000 & 0.7774 \\
graves\_zigzag\_0 & \textbf{1.0000±0.0000} & 0.0000 & 0.7686 \\
graves\_ring\_0 & \textbf{1.0000±0.0000} & 0.0000 & 0.7278 \\
sipu\_a2\_0 & \textbf{0.7555±0.0000} & 0.0000 & 0.4641 \\
wut\_windows\_0 & \textbf{1.0000±0.0000} & 0.0000 & 0.5853 \\
sipu\_a3\_0 & \textbf{0.7434±0.0000} & 0.0000 & 0.1882 \\
\end{longtable}

\begin{figure}[ht]
    \centering
    \includesvg[inkscapelatex=false, width=\textwidth]{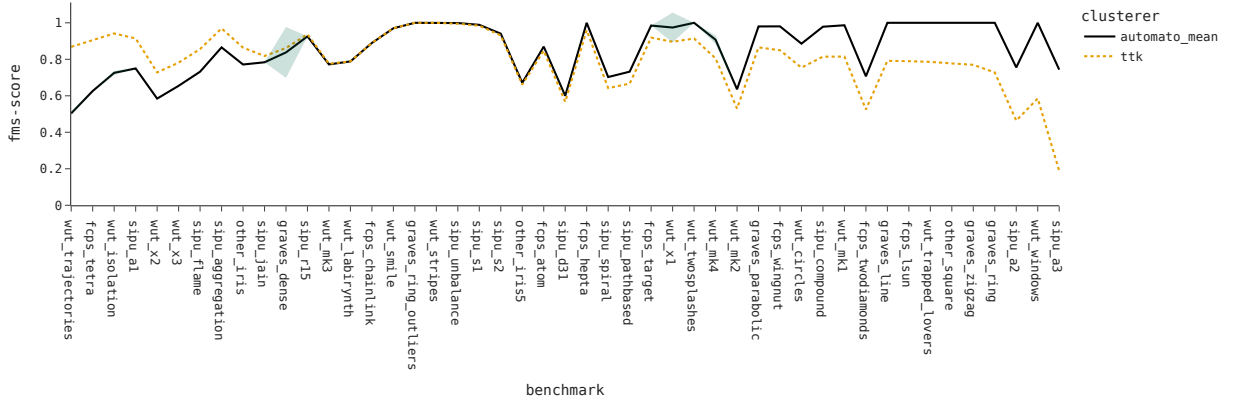}
    \caption{Comparison of \aut and the TTK-algorithm.}
    \label{fig:aut_vs_ttk}
\end{figure}

\begin{longtable}{lrrrr}
\caption{Fowlkes-Mallows scores of AuToMATo vs. hierarchical clustering with complete linkage}
\label{table:aut_vs_linkage_complete} \\
\toprule
Dataset & automato\_mean & linkage\_complete\_max & linkage\_complete\_min \\
\midrule
\endfirsthead
\caption[]{Fowlkes-Mallows scores of AuToMATo vs. hierarchical clustering with complete linkage} \\
\toprule
Dataset & automato\_mean & linkage\_complete\_max & linkage\_complete\_min \\
\midrule
\endhead
\midrule
\multicolumn{5}{r}{Continued on next page} \\
\midrule
\endfoot
\bottomrule
\endlastfoot
sipu\_r15\_2 & 0.4867±0.0000 & \textbf{1.0000} & 0.2256 \\
wut\_trajectories\_0 & 0.5038±0.0107 & \textbf{1.0000} & 0.1706 \\
sipu\_r15\_1 & 0.5436±0.0000 & \textbf{0.8954} & 0.2516 \\
wut\_x3\_1 & 0.6546±0.0000 & \textbf{0.9740} & 0.2004 \\
fcps\_tetra\_0 & 0.6261±0.0000 & \textbf{0.9356} & 0.0651 \\
sipu\_d31\_0 & 0.6001±0.0085 & \textbf{0.8733} & 0.2717 \\
wut\_x3\_0 & 0.5153±0.0000 & \textbf{0.7842} & 0.2477 \\
sipu\_a3\_0 & 0.7434±0.0000 & \textbf{0.8979} & 0.2294 \\
wut\_mk3\_0 & 0.7720±0.0000 & \textbf{0.9207} & 0.0711 \\
wut\_x2\_0 & 0.5846±0.0000 & \textbf{0.7298} & 0.1964 \\
sipu\_a2\_0 & 0.7555±0.0000 & \textbf{0.8992} & 0.2642 \\
sipu\_jain\_0 & 0.7837±0.0000 & \textbf{0.9116} & 0.1288 \\
wut\_circles\_0 & 0.8857±0.0000 & \textbf{1.0000} & 0.1761 \\
sipu\_r15\_0 & 0.9258±0.0000 & \textbf{0.9799} & 0.3372 \\
sipu\_a1\_0 & 0.7499±0.0000 & \textbf{0.8040} & 0.3092 \\
graves\_zigzag\_1 & 0.6720±0.0000 & \textbf{0.7119} & 0.3039 \\
sipu\_aggregation\_0 & 0.8652±0.0000 & \textbf{0.9030} & 0.1246 \\
other\_iris\_0 & 0.7715±0.0000 & \textbf{0.8064} & 0.0680 \\
wut\_x1\_0 & 0.9741±0.0818 & \textbf{1.0000} & 0.2326 \\
sipu\_unbalance\_0 & 0.9986±0.0008 & \textbf{0.9988} & 0.5774 \\
fcps\_hepta\_0 & \textbf{1.0000±0.0000} & \textbf{1.0000} & 0.4321 \\
fcps\_twodiamonds\_0 & \textbf{0.7067±0.0000} & 0.7060 & 0.0916 \\
sipu\_flame\_0 & \textbf{0.7320±0.0000} & 0.7276 & 0.0834 \\
sipu\_compound\_0 & \textbf{0.8616±0.0000} & 0.8472 & 0.1567 \\
sipu\_s1\_0 & \textbf{0.9888±0.0000} & 0.9563 & 0.3672 \\
sipu\_pathbased\_0 & \textbf{0.6517±0.0000} & 0.6022 & 0.1384 \\
sipu\_pathbased\_1 & \textbf{0.7322±0.0000} & 0.6709 & 0.1539 \\
graves\_parabolic\_1 & \textbf{0.6916±0.0000} & 0.6168 & 0.1482 \\
sipu\_compound\_1 & \textbf{0.9786±0.0000} & 0.9023 & 0.1366 \\
sipu\_compound\_4 & \textbf{0.9442±0.0000} & 0.8652 & 0.1408 \\
graves\_dense\_0 & \textbf{0.8377±0.1396} & 0.7584 & 0.3538 \\
wut\_mk1\_0 & \textbf{0.9866±0.0000} & 0.8950 & 0.1591 \\
wut\_smile\_1 & \textbf{0.9701±0.0000} & 0.8697 & 0.3562 \\
graves\_parabolic\_0 & \textbf{0.9802±0.0000} & 0.8610 & 0.1088 \\
wut\_mk2\_0 & \textbf{0.6356±0.0000} & 0.5032 & 0.1096 \\
other\_iris5\_0 & \textbf{0.6712±0.0000} & 0.5305 & 0.0451 \\
fcps\_atom\_0 & \textbf{0.8694±0.0000} & 0.7278 & 0.1364 \\
wut\_smile\_0 & \textbf{0.9681±0.0000} & 0.8206 & 0.3793 \\
sipu\_s2\_0 & \textbf{0.9405±0.0000} & 0.7642 & 0.3114 \\
wut\_mk4\_0 & \textbf{0.9072±0.0234} & 0.7282 & 0.1297 \\
fcps\_target\_0 & \textbf{0.9850±0.0000} & 0.7881 & 0.1934 \\
fcps\_lsun\_0 & \textbf{1.0000±0.0000} & 0.7668 & 0.1377 \\
fcps\_wingnut\_0 & \textbf{0.9805±0.0000} & 0.7406 & 0.0816 \\
graves\_ring\_0 & \textbf{1.0000±0.0000} & 0.7589 & 0.1849 \\
graves\_ring\_outliers\_0 & \textbf{1.0000±0.0000} & 0.7528 & 0.1995 \\
wut\_labirynth\_0 & \textbf{0.7884±0.0000} & 0.4893 & 0.1426 \\
fcps\_chainlink\_0 & \textbf{0.8896±0.0000} & 0.5889 & 0.0919 \\
sipu\_spiral\_0 & \textbf{0.7028±0.0000} & 0.3512 & 0.1424 \\
wut\_isolation\_0 & \textbf{0.7256±0.0113} & 0.3397 & 0.1153 \\
graves\_line\_0 & \textbf{1.0000±0.0000} & 0.5972 & 0.1909 \\
other\_square\_0 & \textbf{1.0000±0.0000} & 0.5846 & 0.1142 \\
graves\_zigzag\_0 & \textbf{1.0000±0.0000} & 0.5505 & 0.2042 \\
wut\_twosplashes\_0 & \textbf{1.0000±0.0000} & 0.5310 & 0.2771 \\
wut\_stripes\_0 & \textbf{1.0000±0.0000} & 0.5136 & 0.0706 \\
wut\_trapped\_lovers\_0 & \textbf{1.0000±0.0000} & 0.4790 & 0.0579 \\
wut\_windows\_0 & \textbf{1.0000±0.0000} & 0.4349 & 0.0781 \\
\end{longtable}

\begin{figure}[ht]
    \centering
    \includesvg[inkscapelatex=false, width=\textwidth]{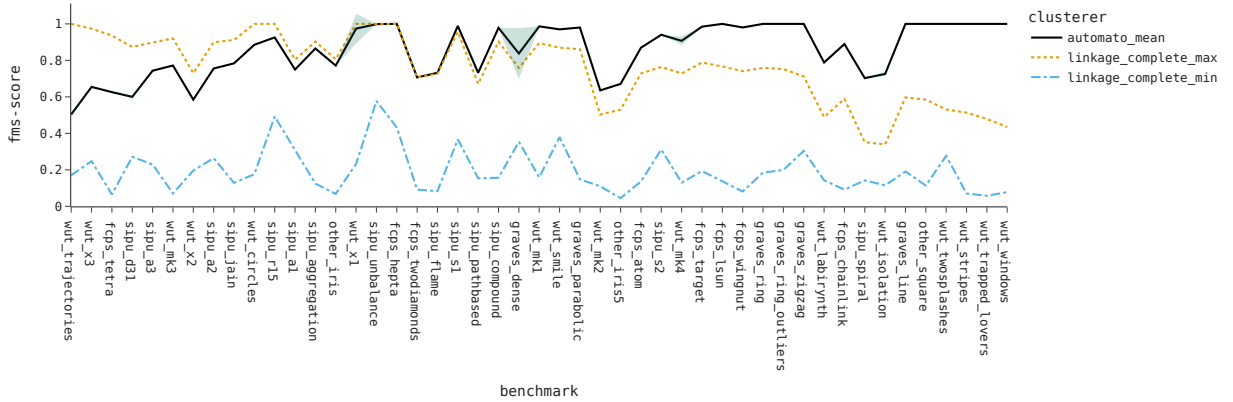}
    \caption{Comparison of \aut and agglomerative clustering with complete linkage.}
    \label{fig:aut_vs_linkage_complete}
\end{figure}

\begin{longtable}{lrrrr}
\caption{Fowlkes-Mallows scores of AuToMATo vs. hierarchical clustering with Ward linkage}
\label{table:aut_vs_linkage_ward} \\
\toprule
Dataset & automato\_mean & linkage\_ward\_max & linkage\_ward\_min \\
\midrule
\endfirsthead
\caption[]{Fowlkes-Mallows scores of AuToMATo vs. hierarchical clustering with Ward linkage} \\
\toprule
Dataset & automato\_mean & linkage\_ward\_max & linkage\_ward\_min \\
\midrule
\endhead
\midrule
\multicolumn{5}{r}{Continued on next page} \\
\midrule
\endfoot
\bottomrule
\endlastfoot
wut\_x3\_0 & 0.5153±0.0000 & \textbf{0.8537} & 0.2203 \\
sipu\_d31\_0 & 0.6001±0.0085 & \textbf{0.9223} & 0.2766 \\
sipu\_a3\_0 & 0.7434±0.0000 & \textbf{0.9377} & 0.2889 \\
sipu\_a2\_0 & 0.7555±0.0000 & \textbf{0.9360} & 0.2653 \\
sipu\_a1\_0 & 0.7499±0.0000 & \textbf{0.9166} & 0.2464 \\
wut\_x2\_0 & 0.5846±0.0000 & \textbf{0.7219} & 0.2076 \\
sipu\_r15\_2 & 0.4867±0.0000 & \textbf{0.5893} & 0.1868 \\
graves\_zigzag\_1 & 0.6720±0.0000 & \textbf{0.7358} & 0.2693 \\
sipu\_r15\_0 & 0.9258±0.0000 & \textbf{0.9832} & 0.4072 \\
sipu\_r15\_1 & 0.5436±0.0000 & \textbf{0.5993} & 0.2082 \\
wut\_x3\_1 & 0.6546±0.0000 & \textbf{0.7090} & 0.1795 \\
wut\_x1\_0 & 0.9741±0.0818 & \textbf{1.0000} & 0.2429 \\
sipu\_unbalance\_0 & 0.9986±0.0008 & \textbf{1.0000} & 0.2063 \\
fcps\_hepta\_0 & \textbf{1.0000±0.0000} & \textbf{1.0000} & 0.4314 \\
sipu\_s1\_0 & \textbf{0.9888±0.0000} & 0.9844 & 0.2453 \\
sipu\_pathbased\_0 & \textbf{0.6517±0.0000} & 0.6251 & 0.1370 \\
sipu\_s2\_0 & \textbf{0.9405±0.0000} & 0.9085 & 0.2177 \\
sipu\_pathbased\_1 & \textbf{0.7322±0.0000} & 0.6844 & 0.1523 \\
fcps\_tetra\_0 & \textbf{0.6261±0.0000} & 0.5622 & 0.0651 \\
graves\_dense\_0 & \textbf{0.8377±0.1396} & 0.7454 & 0.2684 \\
wut\_trajectories\_0 & \textbf{0.5038±0.0107} & 0.3933 & 0.0981 \\
other\_iris\_0 & \textbf{0.7715±0.0000} & 0.6377 & 0.0680 \\
fcps\_atom\_0 & \textbf{0.8694±0.0000} & 0.7272 & 0.1016 \\
graves\_parabolic\_1 & \textbf{0.6916±0.0000} & 0.5260 & 0.1301 \\
fcps\_target\_0 & \textbf{0.9850±0.0000} & 0.7759 & 0.1503 \\
other\_iris5\_0 & \textbf{0.6712±0.0000} & 0.4508 & 0.0451 \\
sipu\_aggregation\_0 & \textbf{0.8652±0.0000} & 0.6064 & 0.1215 \\
sipu\_flame\_0 & \textbf{0.7320±0.0000} & 0.4555 & 0.0819 \\
sipu\_compound\_0 & \textbf{0.8616±0.0000} & 0.5846 & 0.1523 \\
wut\_mk3\_0 & \textbf{0.7720±0.0000} & 0.4911 & 0.0711 \\
fcps\_twodiamonds\_0 & \textbf{0.7067±0.0000} & 0.3967 & 0.0861 \\
sipu\_jain\_0 & \textbf{0.7837±0.0000} & 0.4668 & 0.1201 \\
wut\_mk1\_0 & \textbf{0.9866±0.0000} & 0.6564 & 0.1473 \\
fcps\_lsun\_0 & \textbf{1.0000±0.0000} & 0.6659 & 0.1270 \\
sipu\_compound\_4 & \textbf{0.9442±0.0000} & 0.5793 & 0.1368 \\
sipu\_spiral\_0 & \textbf{0.7028±0.0000} & 0.3100 & 0.1384 \\
wut\_labirynth\_0 & \textbf{0.7884±0.0000} & 0.3749 & 0.0961 \\
sipu\_compound\_1 & \textbf{0.9786±0.0000} & 0.5648 & 0.1327 \\
wut\_twosplashes\_0 & \textbf{1.0000±0.0000} & 0.5817 & 0.2046 \\
wut\_smile\_1 & \textbf{0.9701±0.0000} & 0.5246 & 0.2352 \\
wut\_smile\_0 & \textbf{0.9681±0.0000} & 0.5179 & 0.2505 \\
wut\_mk2\_0 & \textbf{0.6356±0.0000} & 0.1814 & 0.0997 \\
graves\_zigzag\_0 & \textbf{1.0000±0.0000} & 0.5448 & 0.1809 \\
wut\_isolation\_0 & \textbf{0.7256±0.0113} & 0.2309 & 0.0800 \\
graves\_line\_0 & \textbf{1.0000±0.0000} & 0.5045 & 0.1667 \\
wut\_circles\_0 & \textbf{0.8857±0.0000} & 0.3681 & 0.1323 \\
fcps\_chainlink\_0 & \textbf{0.8896±0.0000} & 0.3407 & 0.0894 \\
wut\_mk4\_0 & \textbf{0.9072±0.0234} & 0.3557 & 0.1079 \\
graves\_parabolic\_0 & \textbf{0.9802±0.0000} & 0.4184 & 0.0935 \\
graves\_ring\_0 & \textbf{1.0000±0.0000} & 0.4135 & 0.1427 \\
graves\_ring\_outliers\_0 & \textbf{1.0000±0.0000} & 0.4082 & 0.1515 \\
fcps\_wingnut\_0 & \textbf{0.9805±0.0000} & 0.3229 & 0.0727 \\
other\_square\_0 & \textbf{1.0000±0.0000} & 0.3347 & 0.1007 \\
wut\_windows\_0 & \textbf{1.0000±0.0000} & 0.2833 & 0.0663 \\
wut\_trapped\_lovers\_0 & \textbf{1.0000±0.0000} & 0.2552 & 0.0471 \\
wut\_stripes\_0 & \textbf{1.0000±0.0000} & 0.1922 & 0.0552 \\
\end{longtable}

\begin{figure}[ht]
    \centering
    \includesvg[inkscapelatex=false, width=\textwidth]{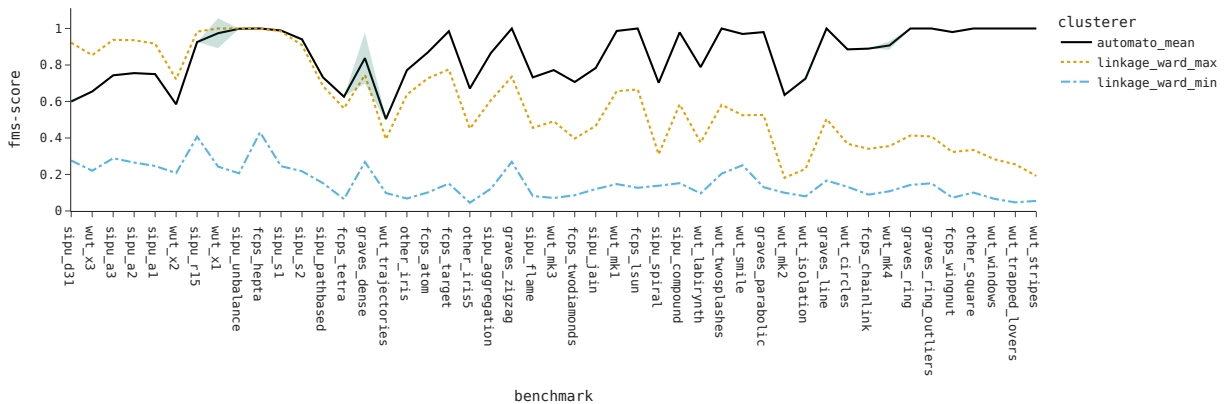}
    \caption{Comparison of \aut and agglomerative clustering with Ward linkage.}
    \label{fig:aut_vs_linkage_ward}
\end{figure}

\begin{longtable}{lrrr}
\caption{Fowlkes-Mallows scores of AuToMATo vs. FINCH}
\label{table:aut_vs_finch} \\
\toprule
Dataset & automato\_mean & automato\_std & finch \\
\midrule
\endfirsthead
\caption[]{Fowlkes-Mallows scores of AuToMATo vs. FINCH} \\
\toprule
Dataset & automato\_mean & automato\_std & finch \\
\midrule
\endhead
\midrule
\multicolumn{4}{r}{Continued on next page} \\
\midrule
\endfoot
\bottomrule
\endlastfoot
wut\_x3\_0 & 0.5153±0.0000 & 0.0000 & \textbf{0.7970} \\
sipu\_d31\_0 & 0.6001±0.0085 & 0.0085 & \textbf{0.8657} \\
sipu\_a3\_0 & 0.7434±0.0000 & 0.0000 & \textbf{0.8306} \\
wut\_x2\_0 & 0.5846±0.0000 & 0.0000 & \textbf{0.6671} \\
wut\_mk3\_0 & 0.7720±0.0000 & 0.0000 & \textbf{0.8503} \\
other\_iris5\_0 & 0.6712±0.0000 & 0.0000 & \textbf{0.7008} \\
sipu\_a2\_0 & 0.7555±0.0000 & 0.0000 & \textbf{0.7635} \\
wut\_x3\_1 & 0.6546±0.0000 & 0.0000 & \textbf{0.6619} \\
sipu\_unbalance\_0 & 0.9986±0.0008 & 0.0008 & \textbf{0.9998} \\
sipu\_r15\_0 & \textbf{0.9258±0.0000} & 0.0000 & 0.9083 \\
wut\_mk1\_0 & \textbf{0.9866±0.0000} & 0.0000 & 0.9655 \\
other\_iris\_0 & \textbf{0.7715±0.0000} & 0.0000 & 0.7477 \\
sipu\_a1\_0 & \textbf{0.7499±0.0000} & 0.0000 & 0.7124 \\
sipu\_r15\_2 & \textbf{0.4867±0.0000} & 0.0000 & 0.4156 \\
graves\_zigzag\_1 & \textbf{0.6720±0.0000} & 0.0000 & 0.5965 \\
graves\_dense\_0 & \textbf{0.8377±0.1396} & 0.1396 & 0.7615 \\
sipu\_r15\_1 & \textbf{0.5436±0.0000} & 0.0000 & 0.4641 \\
sipu\_s1\_0 & \textbf{0.9888±0.0000} & 0.0000 & 0.8728 \\
fcps\_hepta\_0 & \textbf{1.0000±0.0000} & 0.0000 & 0.8794 \\
fcps\_atom\_0 & \textbf{0.8694±0.0000} & 0.0000 & 0.7319 \\
fcps\_tetra\_0 & \textbf{0.6261±0.0000} & 0.0000 & 0.4680 \\
sipu\_s2\_0 & \textbf{0.9405±0.0000} & 0.0000 & 0.7282 \\
wut\_x1\_0 & \textbf{0.9741±0.0818} & 0.0818 & 0.7607 \\
graves\_parabolic\_1 & \textbf{0.6916±0.0000} & 0.0000 & 0.4446 \\
sipu\_flame\_0 & \textbf{0.7320±0.0000} & 0.0000 & 0.4767 \\
sipu\_pathbased\_0 & \textbf{0.6517±0.0000} & 0.0000 & 0.3440 \\
sipu\_compound\_0 & \textbf{0.8616±0.0000} & 0.0000 & 0.5390 \\
sipu\_pathbased\_1 & \textbf{0.7322±0.0000} & 0.0000 & 0.3828 \\
wut\_trajectories\_0 & \textbf{0.5038±0.0107} & 0.0107 & 0.1503 \\
fcps\_lsun\_0 & \textbf{1.0000±0.0000} & 0.0000 & 0.6206 \\
sipu\_compound\_4 & \textbf{0.9442±0.0000} & 0.0000 & 0.5493 \\
sipu\_jain\_0 & \textbf{0.7837±0.0000} & 0.0000 & 0.3824 \\
sipu\_compound\_1 & \textbf{0.9786±0.0000} & 0.0000 & 0.5356 \\
sipu\_spiral\_0 & \textbf{0.7028±0.0000} & 0.0000 & 0.2553 \\
wut\_mk4\_0 & \textbf{0.9072±0.0234} & 0.0234 & 0.4331 \\
wut\_mk2\_0 & \textbf{0.6356±0.0000} & 0.0000 & 0.1478 \\
sipu\_aggregation\_0 & \textbf{0.8652±0.0000} & 0.0000 & 0.3674 \\
fcps\_twodiamonds\_0 & \textbf{0.7067±0.0000} & 0.0000 & 0.1837 \\
wut\_smile\_0 & \textbf{0.9681±0.0000} & 0.0000 & 0.4452 \\
wut\_smile\_1 & \textbf{0.9701±0.0000} & 0.0000 & 0.4181 \\
fcps\_target\_0 & \textbf{0.9850±0.0000} & 0.0000 & 0.4297 \\
wut\_labirynth\_0 & \textbf{0.7884±0.0000} & 0.0000 & 0.2209 \\
wut\_isolation\_0 & \textbf{0.7256±0.0113} & 0.0113 & 0.1440 \\
wut\_twosplashes\_0 & \textbf{1.0000±0.0000} & 0.0000 & 0.4162 \\
graves\_zigzag\_0 & \textbf{1.0000±0.0000} & 0.0000 & 0.4094 \\
graves\_parabolic\_0 & \textbf{0.9802±0.0000} & 0.0000 & 0.3343 \\
graves\_line\_0 & \textbf{1.0000±0.0000} & 0.0000 & 0.3379 \\
fcps\_chainlink\_0 & \textbf{0.8896±0.0000} & 0.0000 & 0.2224 \\
fcps\_wingnut\_0 & \textbf{0.9805±0.0000} & 0.0000 & 0.3110 \\
graves\_ring\_0 & \textbf{1.0000±0.0000} & 0.0000 & 0.2770 \\
wut\_trapped\_lovers\_0 & \textbf{1.0000±0.0000} & 0.0000 & 0.2767 \\
other\_square\_0 & \textbf{1.0000±0.0000} & 0.0000 & 0.2393 \\
graves\_ring\_outliers\_0 & \textbf{1.0000±0.0000} & 0.0000 & 0.2248 \\
wut\_circles\_0 & \textbf{0.8857±0.0000} & 0.0000 & 0.0976 \\
wut\_windows\_0 & \textbf{1.0000±0.0000} & 0.0000 & 0.1166 \\
wut\_stripes\_0 & \textbf{1.0000±0.0000} & 0.0000 & 0.0867 \\
\end{longtable}

\begin{figure}[ht]
    \centering
    \includesvg[inkscapelatex=false, width=\textwidth]{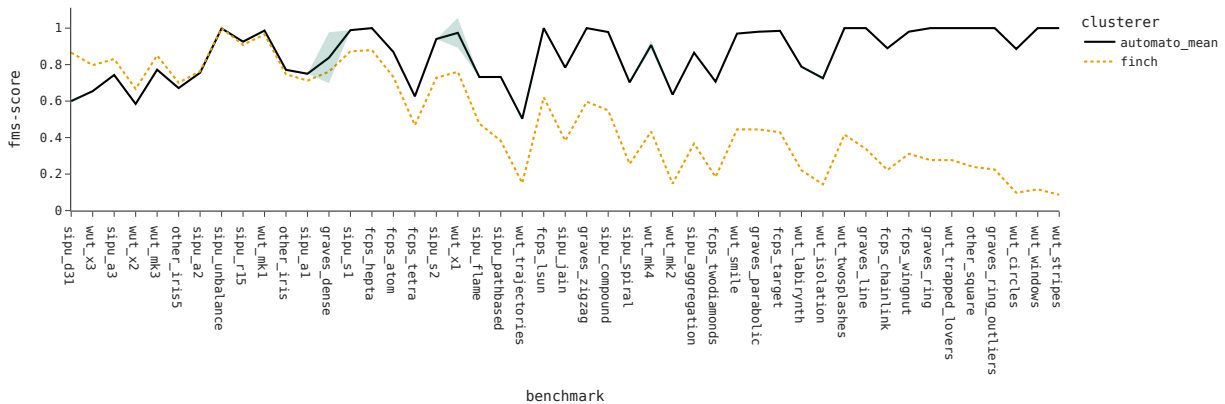}
    \caption{Comparison of \aut and FINCH.}
    \label{fig:aut_vs_finch}
\end{figure}

%%%%%%%%%%%%%%%%%%%%%%%%%%%%%%%%%%%%%%%%%%%%%%%%%%%%%%%%%%%%%%%%%%%%%%%%%%%%%%%%%%%%%%%%%%%%%%%%

\end{document}